\documentclass[10pt,twocolumn,letterpaper]{article}

\usepackage[pagenumbers]{iccv} 

\definecolor{iccvblue}{rgb}{0.21,0.49,0.74}
\usepackage[pagebackref,breaklinks,colorlinks,allcolors=iccvblue]{hyperref}
\usepackage{tabu}
\usepackage{dsfont}

\newcommand{\myparagraph}[1]{\vspace{2pt}\noindent{\bf #1}}

\def\paperID{8424} 
\def\confName{ICCV}
\def\confYear{2025}

\title{PASTA: Part-Aware Sketch-to-3D Shape Generation with Text-Aligned Prior}

\author{Seunggwan Lee\textsuperscript{1}\hspace{0.3em}
Hwanhee Jung\textsuperscript{1}\hspace{0.4em}
Byoungsoo Koh\textsuperscript{2}\hspace{0.4em}
Qixing Huang\textsuperscript{3}\hspace{0.4em}
Sangho Yoon\textsuperscript{4}\hspace{0.4em}
Sangpil Kim\textsuperscript{1}\footnotemark[1] \\
\\
\textsuperscript{1}Korea University \quad
\textsuperscript{2}KOCCA \quad
\textsuperscript{3}The University of Texas at Austin \quad
\textsuperscript{4}KAIST\\
{\tt\small \{gwan7801, hwanyo14, spk7\}@korea.ac.kr} \quad
{\tt\small kbs0753@gmail.com} \\
{\tt\small huangqx@cs.utexas.edu} \quad
{\tt\small sangho@kaist.ac.kr}
}
\begin{document}
\twocolumn[{%
\renewcommand\twocolumn[1][]{#1}%
\vspace{-5.5em}
\maketitle
    \begin{center}
    \centering 
    \includegraphics[width=1.0\linewidth]{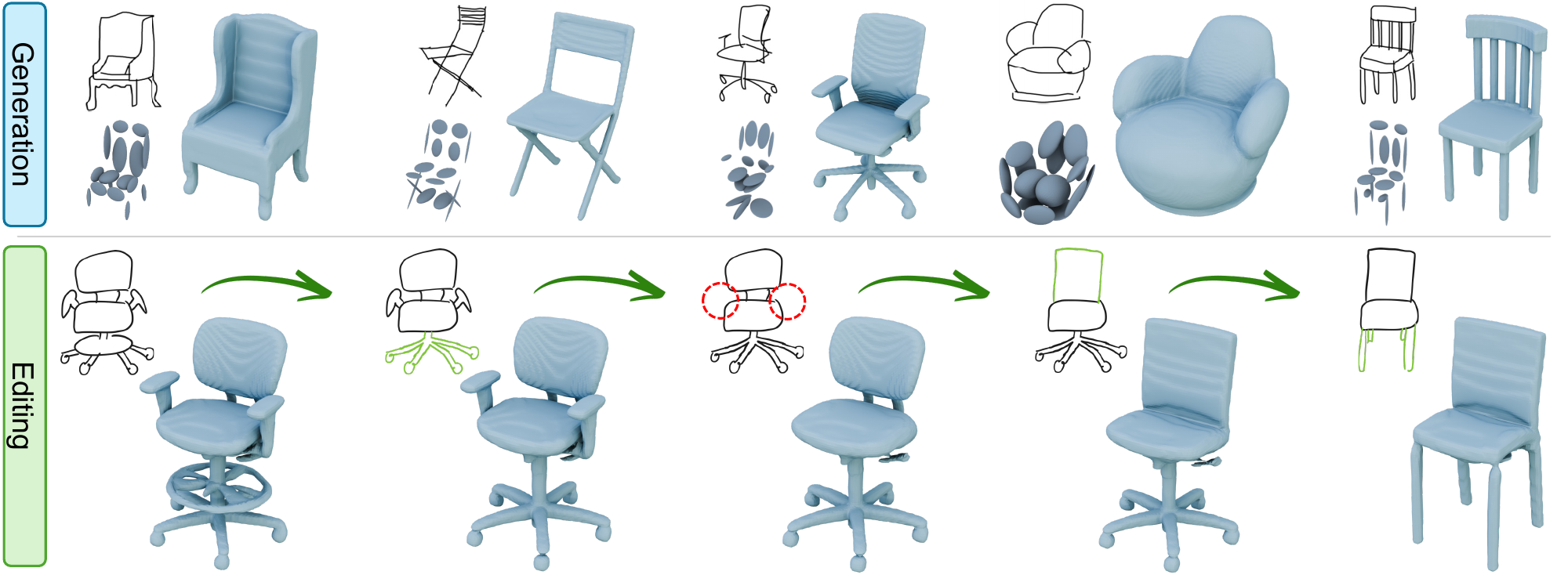}
    \captionof{figure}{
    \textbf{PASTA} enables 3D shape generation and editing from hand-drawn sketches. It processes a sketch input using Gaussian mixture models (GMMs) for part-level decomposition, generating a corresponding 3D mesh. The framework also supports localized modifications, such as adding, deleting, or transforming parts, for precise user interactive editing.
    }
    \label{fig:teaser}
    \end{center}
}]
\maketitle
\begin{abstract}
A fundamental challenge in conditional 3D shape generation is to minimize the information loss and maximize the intention of user input. Existing approaches have predominantly focused on two types of isolated conditional signals, i.e., user sketches and text descriptions, each of which does not offer flexible control of the generated shape. In this paper, we introduce \textbf{PASTA}, the flexible approach that seamlessly integrates a user sketch and a text description for 3D shape generation. The key idea is to use 
text embeddings from a vision-language model to enrich the semantic representation of sketches. Specifically, these text-derived priors specify the part components of the object, compensating for missing visual cues from ambiguous sketches.
In addition, we introduce ISG-Net which employs two types of graph convolutional networks: IndivGCN, which processes fine-grained details, and PartGCN, which aggregates these details into parts and refines the structure of objects. Extensive experiments demonstrate that PASTA outperforms existing methods in part-level editing and achieves state-of-the-art results in sketch-to-3D shape generation.
\vspace{-3em}
\renewcommand{\thefootnote}{}
\footnotetext{\hspace{-1.5em}\textsuperscript{*}Corresponding author}
\end{abstract}   
\section{Introduction}
\label{sec:intro}

Conditional 3D shape generation, which generates a 3D shape from user constraints, has many applications in design, gaming, and virtual reality. So far, existing approaches have predominantly focused on two types of conditional signals, namely user sketches~\cite{zhong2020towards, zhong2022study, li2020sketch2cad, li2022free2cad, lun20173d, li2018robust} and text descriptions~\cite{DBLP:conf/cvpr/LiuWQF22,DBLP:journals/tog/ZhangTNW23,DBLP:conf/cvpr/ChengLTSG23,sanghi2023clip, nichol2022point}. Most efforts focus on building machine learning models to achieve improved performance under these two conditional generation tasks. In this paper, we look at the critical question of what are the most effective and informative conditional signals for 3D shape generation? Such conditional signals should be concise, easy to specify, and offer accurate control of the shape to be generated. Unfortunately, both sketch-based and text-based conditions exhibit fundamental limitations. Text descriptions are certainly concise and offer shape semantics, but do not provide precise control of shape geometries. User sketches, on the other hand, offer better control on shape geometries, but they suffer from limited semantics and ambiguities in a single 2D sketch, c.f.,~\cite{binninger2024sens, guillard2021sketch2mesh, bandyopadhyay2024doodle} . In this paper, we study how to develop a new conditional 3D shape generation approach by seamlessly integrating user sketches and text descriptions. 

To this end, we propose a novel framework called PASTA that effectively integrates complementary information to generate 3D models from sketches. 
Specifically, to compensate for semantic cues that the visual backbone might overlook, we leverage text embeddings extracted from the input sketches. 
These text embeddings, which are grounded in the rich knowledge of a pre-trained vision-language model (VLM)~\cite{liu2023visual}, provide detailed descriptions of object components, such as the number of legs on a chair or the presence of an armrest, that enhance the information provided by the visual backbone. By integrating both text and visual information, PASTA can more precisely understand object composition and semantics, ultimately generating accurate 3D models.

Similar to previous studies~\cite{binninger2024sens, bandyopadhyay2024doodle}, our model represents 3D shapes via decomposed Gaussian mixture models(GMMs) through SPAGHETTI shape decoder~\cite{hertz2022spaghetti}, making interactions between GMMs critically important.
Accordingly, along with text priors, we introduce the Integrated Structure-Graph Network (ISG-Net), a specialized refinement module that employs two types of graph convolutional networks (GCNs) to process the sketch. 
The first network, IndivGCN, focuses on processing the details of each individual GMM, capturing subtle features within the sketch.
The second network, PartGCN, aggregates these fine details into parts and refines the structure of the object, improving the consistency
and spatial relationships between the different parts.
Together, these components enable a more comprehensive structural understanding of the sketch, facilitating part-level editing with superior accuracy and flexibility compared to existing methods. 

To evaluate the robustness of PASTA, We also conducted an additional experiment to extend the domain to real-world images.
Our comprehensive experiments demonstrate that PASTA achieves state-of-the-art performance in qualitative and quantitative evaluations, generating more complete and structurally sound 3D models than prior approaches. 
\begin{figure*}[t]
    \centering
    \includegraphics[width=1.0\textwidth]{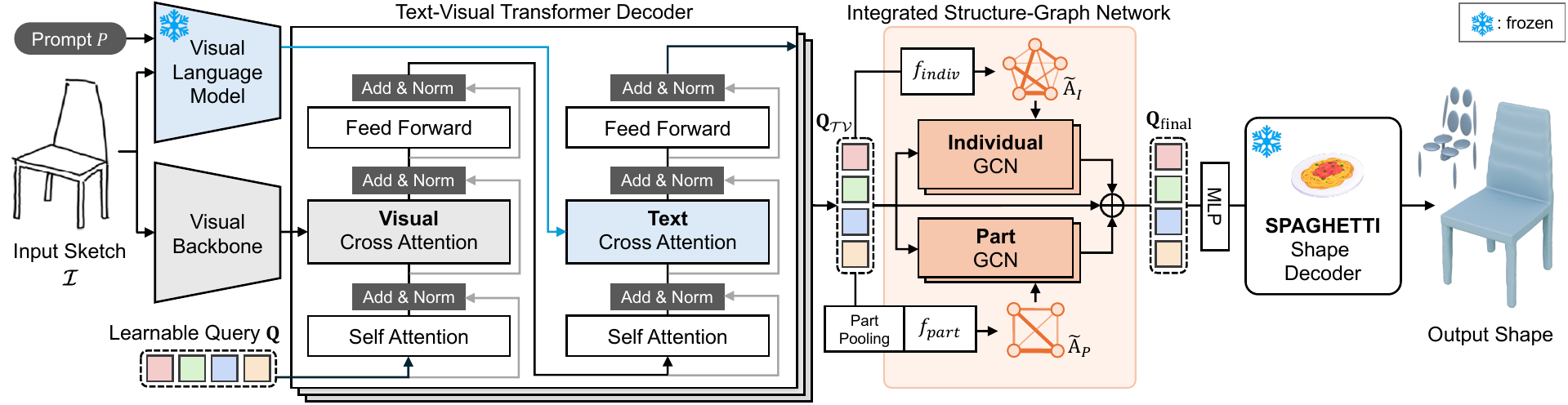}
    \caption{
    \textbf{Overview of PASTA.} Our framework enhances sketch-based 3D shape generation by integrating visual embeddings and text-aligned priors. A visual backbone and vision-language model (VLM) extract meaningful features from an input sketch, which are then processed by a Text-Visual Transformer Decoder with learnable queries. To refine structural details, we introduce ISG-Net, which consists of IndivGCN for fine-grained feature processing and PartGCN for aggregating part-level information. The output features are fed into the SPAGHETTI shape decoder~\cite{hertz2022spaghetti}, producing a more complete and structurally accurate 3D model.
    }
    \label{fig:example_figure}
    \vspace{-1.em}
\end{figure*}
The summary of our contributions is as follows:

\begin{itemize} 
\item We propose PASTA, a novel approach for sketch-to-3D shape generation that integrates visual and text conditions from VLM to compensate for missing visual cues, enhancing the comprehension of object components in the provided user sketch.
\item We introduce ISG-Net, a specialized refinement module incorporating two types of GCNs: IndivGCN for capturing fine-grained details and PartGCN for ensuring structural integrity at the part-level. 
\item Achieving state-of-the-art performance, PASTA outperforms existing methods in qualitative and quantitative evaluations, generating structurally accurate 3D models.
\item Additionally, the scalability of PASTA is demonstrated, showing its ability to extend to real-world images and effectively handle realistic visual data beyond sketches.
\end{itemize}
\section{Related Work}
\label{sec:relatedwork}

\myparagraph{3D Shape Representation.}
Unlike 2D grid images with RGB channels, 3D data can be represented in a variety of ways.
Voxel-based 3D shape generation methods~\cite{wu2016learning, liu2017interactive, smith2017improved} have been widely explored; however, they incur significant computational costs and suffer from limited expressive power at low resolutions. 
Another approach employs point cloud representations~\cite{li2018point, achlioptas2018learning, luo2021diffusion,zhou20213d}, which offer a sparser and more lightweight alternative, albeit being unstructured and lacking explicit connectivity information. 
To address these limitations, numerous studies have focused on mesh-based 3D shape generation~\cite{zhang2021sketch2model, gao2022get3d, wang2018pixel2mesh}. 
However, mesh-based approaches often necessitate complex operations to manage topological variations and high-genus structures. 
More recently, neural implicit representations such as signed distance functions and occupancy fields have garnered considerable attention due to their flexibility and ability to model shapes at arbitrary resolutions with compelling performance~\cite{chen2019learning, mittal2022autosdf, cheng2023sdfusion, zheng2023locally, kleineberg2020adversarial}. 
In this work, we adopt an implicit modeling strategy~\cite{hertz2022spaghetti} based on a Gaussian mixture model (GMM), which enables flexible shape representation and part-level editing.

\myparagraph{Sketch-based 3D Shape Generation.}
Sketch-based 3D shape generation seeks to reconstruct or generate 3D shapes from 2D sketches, bridging the gap between abstract drawings and detailed models. 
Prior works have focused on single-view~\cite{zhong2020towards, zhong2022study, li2020sketch2cad, li2022free2cad} or multi-view sketches~\cite{lun20173d, li2018robust}, yet these methods often fail to recover intricate details due to the oversimplification inherent in the sketches, inconsistencies across different views, complicating the reconstruction process. 
Recent view-aware approaches~\cite{zheng2023locally, zhou2023ga} have made notable improvements by incorporating viewpoint-specific information. In addition, user interactive methods~\cite{guillard2021sketch2mesh, binninger2024sens, bandyopadhyay2024doodle} have similarly advanced the field by enabling user-guided editing.
Despite these advancements, fully reconstructing the intended shape from sketches remains challenging.
Therefore, in this paper, we propose a method that leverages part-level representations~\cite{hertz2022spaghetti} alongside text priors and graph structures to reinforce semantic representation and inter-part relationships, thereby generating more accurate and consistent 3D models.

\myparagraph{Vision-Language Pre-trained Models.}
In computer vision, pre-trained vision-language models (VLMs) have exerted a profound impact by integrating textual modalities. 
Pre-trained on large-scale image-text pairs~\cite{zhou2022learning, chen2020uniter, long2022vision}, these models have been successfully applied to a variety of downstream tasks, including visual question answering, text-to-image generation, and image captioning. 
Notably, models such as CLIP~\cite{radford2021learning} and ALIGN~\cite{jia2021scaling} utilize contrastive learning to acquire cross-modal representations, thereby enabling the understanding of visual concepts in zero-shot scenarios. 
More recently, research has leveraged the rich knowledge embedded within large language models (LLMs)~\cite{chiang2023vicuna, chung2024scaling, anil2023palm, brown2020language} to enhance visual context comprehension and image processing~\cite{awadalla2023openflamingo, liu2023visual, li2023blip}. 
This paradigm has further extended to 3D vision tasks, such as 3D scene understanding~\cite{huang2023clip2point, hong20233d, qi2024shapellm, roh2024catsplat, roh2023functional, roh2024edge}. 
In our work, we exploit VLMs to accurately recognize object components from sketch images and to strengthen the modeling of their relationships, thereby addressing the inherent limitations of abstract sketches.

\section{Method}
\label{sec:method}
Our proposed method, PASTA, aims to generate 3D shapes from hand-drawn sketches.
To enable more flexible shape modeling, we encode sketches in the latent space of the part-level implicit shape decoder~\cite{hertz2022spaghetti}.
Consequently, it is crucial to map the information from the sketches accurately to aligned latent vectors, which are then represented as Gaussian mixture models (GMMs) for generating the 3D shapes.
To achieve this, we utilize text-aligned priors and graph-based processing.
In Fig.~\ref{fig:example_figure}, we provide an overview of the framework.
Given a sketch, a visual backbone~\cite{oquab2023dinov2} processes the input sketch image, while a vision-language model (VLM)~\cite{liu2023visual} captures semantic priors from a sketch.
A Text-Vision Transformer Decoder then processes these features using learnable queries to fuse and refine visual and text information.
To ensure structural and part-level consistency, we introduce the Integrated Structure-Graph Network (ISG-Net), which comprises an Individual Graph Convolutional Network (IndivGCN) for local feature refinement and a Part Graph Convolutional Network (PartGCN) for maintaining part-level coherence.
The refined features are finally fed into the shape decoder~\cite{hertz2022spaghetti} to generate a complete 3D shape that reflects the sketch and its semantic attributes. Section~\ref{sec:text_prior} details the Text-Vision Transformer Decoder, while Section~\ref{sec:gs_net} discusses the ISG-Net module.
\subsection{Text-Visual Transformer Decoder}\label{sec:text_prior}
Due to the ambiguous and oversimplified nature of the sketches, inaccurate shape generation and part omission often occur.
To address these challenges, we introduce a text-aligned prior that integrates visual and text condition into the 3D shape generation process. 
This prior is extracted from vision-language model (VLM)~\cite{liu2023visual} that analyze the overall composition of the sketch and, through a text-attention mechanism, integrate visual cues to enrich the highly abstract characteristics of the sketch. 
The integration of these components ensures that the generated shapes are coherently aligned with semantic attributes derived from text embeddings, even when the sketches do not fully convey the intended details. 
Consequently, our approach improves the ability of the network to perceive the composition of the sketch, resulting in a more robust understanding.

\myparagraph{Vision-Language Embedding.}  
Starting with an input sketch, we extract feature representations from distinct modalities using a dual-branch approach.
Firstly, the visual backbone~\cite{oquab2023dinov2} produces a visual embedding $\mathcal{V} = f_{\text{vis}}(\mathcal{I})$, where $\mathcal{I}$ denotes the input sketch and $f_{\text{vis}}(\cdot)$ denotes the forward pass of the visual backbone encoder. 
However, relying solely on this visual information is insufficient to capture the semantic nuances from a simple sketch, leading to suboptimal reconstruction performance.

\begin{figure}[t]
    \centering
    \vspace{-1.em}
    \includegraphics[width=\linewidth]{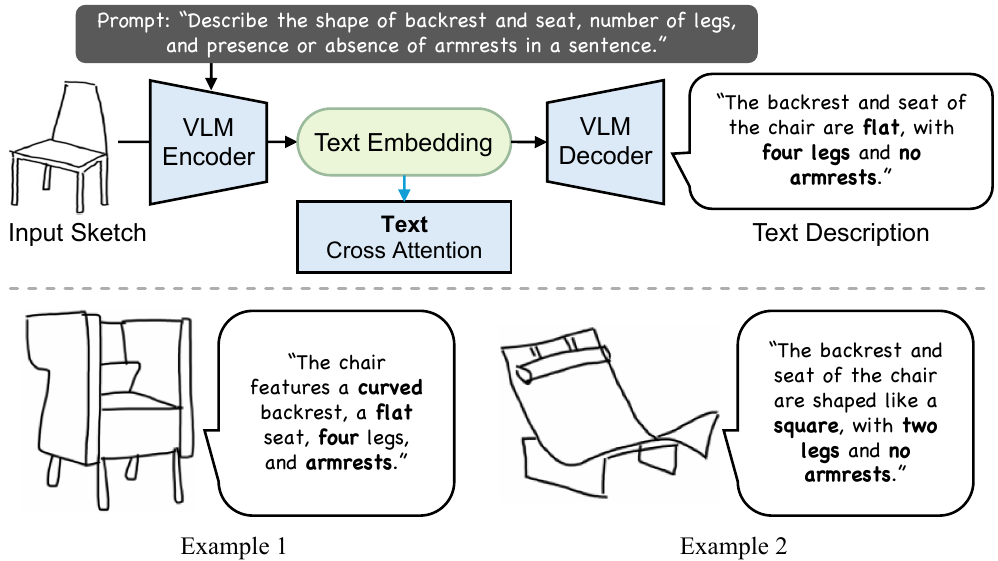}
    \vspace{-1.5em}
    \caption{
    An example of the prompts we used for chair sketches and their various corresponding text descriptions from the VLM~\cite{liu2023visual}.
    The figure illustrates how the generated text descriptions emphasize key semantic details of the sketch.
    }
    \label{fig:text}
    \vspace{-1.em}
\end{figure}

Therefore, we leverage semantically meaningful text embeddings obtained from the pre-trained VLM~\cite{liu2023visual}, thereby exploiting the knowledge encapsulated within this model.
Text embeddings are computed as $\mathcal{T} = f_{\text{vlm}}(\mathcal{I}, P)$, where $P$ denotes the input prompt and $f_{\text{vlm}}(\cdot)$ represents the VLM text and image encoder. 
As illustrated in Fig.~\ref{fig:text}, the obtained $\mathcal{T}$ represents the embedding immediately preceding the conversion to the output text description and contains information regarding the overall composition of the sketch (\eg, the shape of the backrest and seat, the number of legs and the absence of armrests).

\myparagraph{Text-Conditioned Query Learning.}
For accurate 3D shape generation, we employ a Text-Visual Transformer Decoder to integrate visual embeddings $ \mathcal{V}$ and text embeddings $\mathcal{T}$. 
As the shape decoder~\cite{hertz2022spaghetti} expects $N$ latent vectors as input to form $N$ GMMs, we define learnable queries $\mathbf{Q} = [\mathbf{q}_1, \mathbf{q}_2, \dots, \mathbf{q}_N] \in \mathbb{R}^{N \times d}$, where $d$ denotes the dimensionality of each query.
Initially, queries are processed by a self-attention mechanism to capture their internal dependencies. 
For the attention operation, $\mathbf{Q}$ is linearly projected via learnable weight matrices $W_Q$, $W_K$, and $W_V$, which map $\mathbf{Q}$ to query, key, and value vectors, respectively, which is expressed as:
\begin{equation} \label{eq:QKV}
Q = W_Q^T \cdot \mathbf{Q}, \quad K = W_K^T \cdot \mathbf{Q}, \quad V = W_V^T \cdot \mathbf{Q},
\end{equation}
\begin{equation} \label{eq:self_attn}
\mathbf{Q} = Attn(Q, K, V) = softmax(\frac{Q \cdot K^T}{\sqrt{d}}) V.
\end{equation}
The resulting queries are added to the residuals and normalized before being subjected to cross-attention with the visual embeddings $\mathcal{V}$.
Through visual cross-attention, each of the $N$ queries acquires visual embeddings extracted from sketch, a process that can be expressed as:
\begin{equation} \label{eq:cross_attn1}
\mathbf{Q}_\mathcal{V} = Attn\left(W_Q^T \cdot \mathbf{Q}, W_K^T \cdot \mathcal{V}, W_V^T \cdot \mathcal{V}\right).
\end{equation}
During training, the learnable queries are optimized to retrieve the most representative visual cues for each vector $\mathbf{q}$.
Next, the queries $\mathbf{Q}_\mathcal{V}$ enter another round of self-attention and feed-forward transformation, further refining their relationships. 
Subsequently, queries are integrated with text embeddings $\mathcal{T}$ obtained from the VLM through a text cross-attention mechanism:
\begin{equation} \label{eq:cross_attn2}
\mathbf{Q}_\mathcal{TV} = Attn\left(W_Q^T \cdot \mathbf{Q}_\mathcal{V}, W_K^T \cdot \mathcal{T}, W_V^T \cdot \mathcal{T}\right).
\end{equation}
This operation ensures that queries $\mathbf{Q}_\mathcal{TV}$ are conditioned not only on visual information but also on the semantic information embedded in the text representation, effectively integrating both modalities.
By incorporating text embeddings, the model enriches the visual queries with complementary linguistic context, allowing for a more comprehensive understanding of object.
Finally, a feed-forward layer processes the updated queries $\mathbf{Q}_\mathcal{TV}$ before passing them to the next iteration. 
This entire process is repeated 12 times, each iteration allowing the queries to progressively capture fine-grained semantic details that the visual backbone might have otherwise overlooked by integrating information from text and visual embeddings.

\subsection{ISG-Net}\label{sec:gs_net}
To enhance the structural coherence and geometric accuracy of the generated 3D shapes, we introduce ISG-Net, a GCN-based module that refines the queries $\mathbf{Q}_\mathcal{TV}$ through relational reasoning. 
ISG-Net consists of two key components: IndivGCN, which refines fine-grained local features, and PartGCN, which aggregates part-level structural information to enforce global consistency. 
By jointly leveraging these two networks, ISG-Net encodes both local geometric details and high-level structural relationships, ensuring a faithful generation of the target 3D shape.
\subsubsection{IndivGCN: Fine-Grained Feature Processing}
The IndivGCN module considers the relationships among individual queries to integrate features from neighboring nodes. 
For graph-based operations, obtaining an adjacency matrix is essential; therefore, we introduce an individual adjacency predictor $f_{\text{indiv}}$ composed of a multilayer perceptron (MLP) to define the interquery relationships. 
As illustrated in Fig.~\ref{fig:gcn} (a), the predicted $ \mathbf{Q}_{I} = f_{\text{indiv}}(\mathbf{Q}_{\mathcal{TV}})$ is used to compute the individual adjacency matrix $\tilde{\mathbf{A}}_{I} = \mathbf{Q}^{T}_{I}\mathbf{Q}_{I} \in \mathbb{R}^{N \times N}$,
which is supervised by a pseudo-ground truth $\mathbf{A}_{I}$.  
The loss $ \mathcal{L}_{\text{indiv}}$ is computed using the mean squared error (MSE) as follows:
\vspace{-5pt}
\begin{equation} \label{eq:loss_indiv}
\mathcal{L}_{\text{indiv}} = \frac{1}{N^2} \sum_{i=1}^{N} \sum_{j=1}^{N} \left\| \mathbf{\tilde{A}}_I(i,j) - \mathbf{A}_I(i,j) \right\|_2^2.
\end{equation}
Regarding the pseudo-ground truth, $\mathbf{A}_{I}$ is derived based on the positions $\mu \in \mathbb{R}^{N \times 3}$ of the $N$ GMMs. 
The 3D shape is converted into GMMs by inversion of SPAGHETTI~\cite{hertz2022spaghetti}, and the distances between all Gaussian means $\mu$ are computed. 
Note that these distances are normalized to the interval between 0 and 1, then inverted, as smaller distances correspond to stronger connections.
Through this, $f_{\text{indiv}}$ is optimized to predict the distances between the actual GMM centers, and the derived adjacency matrix is then employed to perform graph convolution as follows:
\setlength{\abovedisplayskip}{0.5em} 
\setlength{\belowdisplayskip}{0.5em} 
\begin{equation} \label{eq:gsmet4}
\mathbf{Q}_{\text{indiv}} = \sigma(\tilde{\mathbf{A}}_{I} \mathbf{Q}_{\mathcal{TV}} \mathbf{W}_{I}),
\end{equation}
where \( \sigma \) is a non-linear activation function and $\mathbf{W}_{I}$ is the learnable weight matrix for graph convolution. 
This formulation allows the network to dynamically learn relationships, refining feature representations of individual nodes.

\clearpage
\subsubsection{PartGCN: Structural aggregate at the part-level}
\begin{figure}[t]
    \centering
    \vspace{-1.em}
    \includegraphics[width=\linewidth]{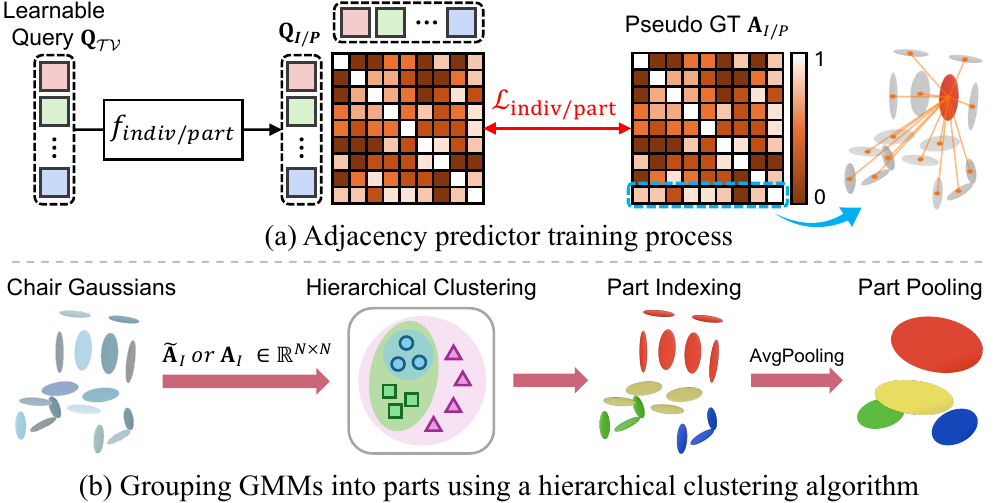}
    \vspace{-1.5em}
    \caption{
    Detailed illustration of ISG-Net. (a) Shows the training process of the adjacency predictor, which learns the relationships between queries. (b) Demonstrates the process of grouping GMMs into parts using a hierarchical clustering algorithm to better capture structural dependencies.
    }
    \label{fig:gcn}
    \vspace{-1.em}
\end{figure}

To further refine the details through IndivGCN and enforce structural consistency, we perform graph convolution by grouping GMMs into subsets at the part-level. 
First, hierarchical clustering~\cite{zhao2005hierarchical} is used to produce indices that assign the $N$ queries to the $K$ parts ($K < N$), with further details provided in the supplementary material. 
The indices obtained are used to perform an average pooling operation on the queries, resulting in part-level queries
$\mathbf{Q}_P \in \mathbb{R}^{K \times d}$, which in turn are used by $f_{\text{part}}$ to predict $\tilde{\mathbf{A}}_{P} \in \mathbb{R}^{K \times K}$. 
The loss $\mathcal{L}_{\text{part}}$ is also calculated using MSE as follows:
\vspace{-0.2em}
\begin{equation} \label{eq:loss_part}
\mathcal{L}_{\text{part}} = \frac{1}{K^2} \sum_{i=1}^{K} \sum_{j=1}^{K} \left\| \mathbf{\tilde{A}}_P(i,j) - \mathbf{A}_P(i,j) \right\|_2^2,
\end{equation}
\vspace{-0.2em}
where $\mathbf{A}_{P}$ denotes the adjacency matrix of the pseudo-ground truth part.
As shown in Fig.~\ref{fig:gcn} (b), it is constructed based on the part coordinates obtained by applying clustering on the distances of the GMMs, similar to $\mathbf{A}_{I}$. With the queries $\mathbf{Q}_\mathcal{TV}$ and the predicted $\tilde{\mathbf{A}}_P$, we perform the following operations:
\vspace{-0.4em}
\begin{equation} \label{eq:gsmet5}
\mathbf{Q}_{\text{part}} = \sigma(\tilde{\mathbf{A}}_{P} \mathbf{Q}_{\mathcal{TV}} \mathbf{W}_{P}).
\end{equation}
After obtaining the part-level representations, the features are unpooled to match the size of the individual queries, ensuring compatibility with IndivGCN. 
The final representation is obtained by fusing the output from IndivGCN and PartGCN with the residual connection:
\begin{equation} \label{eq:gsmet6}
\mathbf{Q}_{\text{final}} = norm(\alpha \mathbf{Q}_{\text{indiv}} + (1 - \alpha) \mathbf{Q}_{\text{part}} + \mathbf{Q}_\mathcal{TV}),
\end{equation}
where $\alpha$ is a weighting factor that balances the impact of two GCNs. 
The result is subsequently normalized to produce $\mathbf{Q}_{\text{final}}$, which is processed by an MLP to infer latent vectors $\tilde{\mathbf{z}}$.
These vectors are then passed through the shape decoder, which predicts occupancy fields and generates a 3D mesh via the marching cubes algorithm~\cite{lorensen1998marching}.

\subsubsection{Loss Function}
To leverage the part-level implicit shape decoder~\cite{hertz2022spaghetti}, that is crucial that the vectors predicted by our network are accurately aligned with the latent space of the pre-trained shape decoder. To ensure that the predicted vectors $\tilde{\mathbf{z}}$ remain close to the ground truth latent vectors $\mathbf{z}$ in the feature space, we optimize the network using the following loss function:
\vspace{-0.2em}
\begin{equation} \label{eq:loss_full}
\mathcal{L}_{\text{align}} = \frac{1}{N} \sum_{i=1}^{N} \| \tilde{\mathbf{z}}_i - \mathbf{z}_i \|_1,
\end{equation}
\vspace{-0.2em}
where $\mathbf{z}$ represents the latent vectors extracted from the 3D shape using the inversion of SPAGHETTI~\cite{hertz2022spaghetti}. The total loss function $\mathcal{L}$ is defined as the sum of three components:
\begin{equation} \label{eq:loss_final}
\mathcal{L} = \lambda_{\text{align}} \mathcal{L}_{\text{align}} + 
\lambda_{\text{indiv}} \mathcal{L}_{\text{indiv}} + \lambda_{\text{part}} \mathcal{L}_{\text{part}}.
\end{equation}
In this equation, $\lambda$ denotes the weighting factors for each loss term. By jointly optimizing all three losses, PASTA achieves structural coherence in 3D shape generation.
\section{Experiments}
\label{sec:experiments}
\subsection{Experimental Setup}
\vspace{-.4em}
\myparagraph{Datasets.}
We employ a subset of the ShapeNet dataset ~\cite{chang2015shapenet} along with the corresponding sketches. 
Specifically, the chair dataset comprises 6,755 shapes, the airplane dataset contains 1,575, and the lamp dataset includes 733.
Each shape is normalized to a unit cube scale and rendered from 6 viewpoints. 
Abstract sketches are obtained using CLIPasso~\cite{vinker2022clipasso} from these rendered images.
For the chair category, we additionally leverage non-photo-realistic renderings~\cite{chan2022learning}, which map photographs into line drawings.
\begin{figure*}[t]
    \centering
    \includegraphics[width=1.0\textwidth]{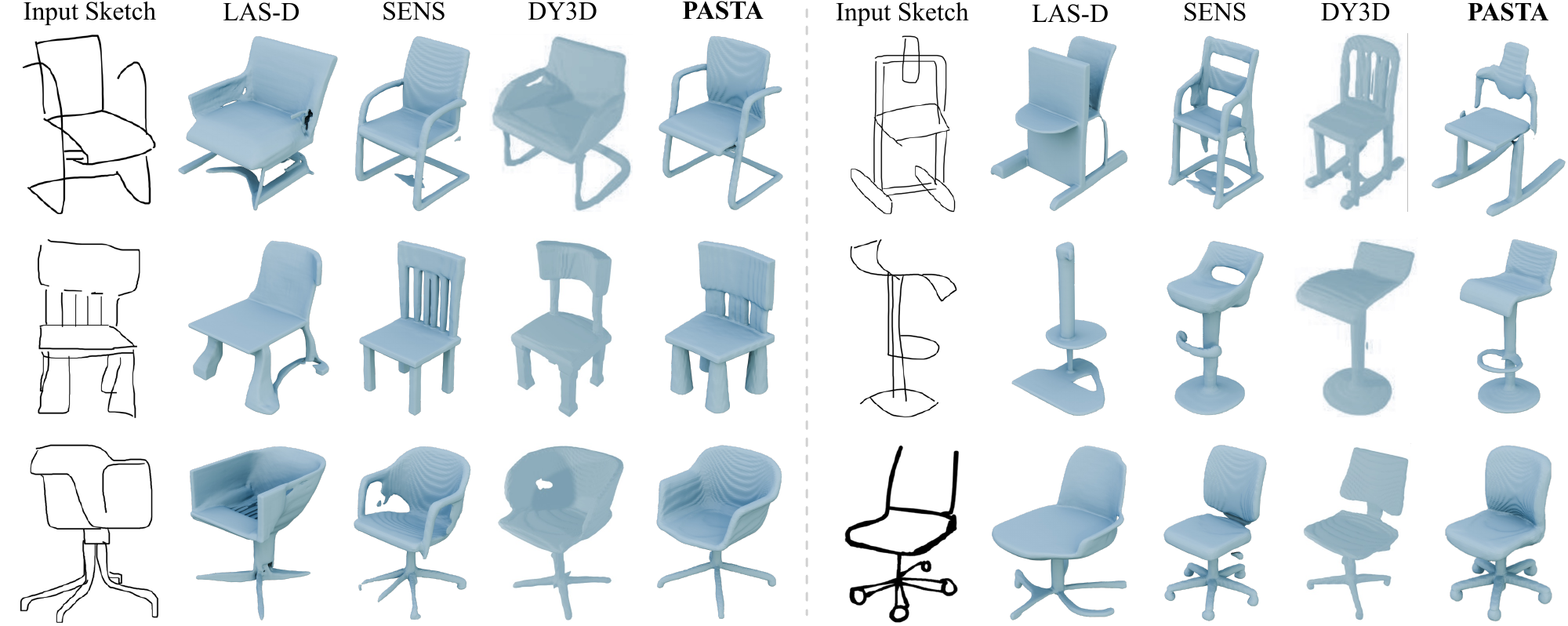}
    \caption{Qualitative comparison of chair reconstructions. The figure demonstrates the effectiveness of our methods in generating high-quality 3D chair models, accurately capturing intricate details such as armrests, backrests, and leg structures, while also preserving overall global structures, compared to LAS-D~\cite{zheng2023locally}, SENS~\cite{binninger2024sens}, and DY3D~\cite{bandyopadhyay2024doodle}.}
    \label{fig:Qualitative}
    \vspace{-1em}
\end{figure*}

For evaluation, we conduct two experiments for the chair category.
First, we utilize the AmateurSketch-3D~\cite{qi2021toward} dataset containing 1,000 chair shapes along with corresponding hand-drawn sketches from three different viewpoints.
Second, we evaluate 500 chair shapes on the expert-drawn sketch dataset, ProSketch-3D\cite{zhong2020towards}, which provides more detailed representations. 
For the airplane and lamp categories, sketches corresponding to 200 airplane shapes and 100 lamp shapes are obtained with CLIPasso.
Following the protocols adopted in prior work~\cite{park2019deepsdf, binninger2024sens, bandyopadhyay2024doodle}, the latent vectors used during training are derived from 3D shapes through the inversion functionality provided by SPAGHETTI~\cite{hertz2022spaghetti}.

\myparagraph{Evaluation Metrics.}
We evaluate the quality of the 3D shapes generated using three metrics, with lower values indicating better results.
First, Chamfer Distance (CD) measures the average distance between the corresponding points in two-point clouds. 
Similarly, Earth Mover Distance (EMD) quantifies the minimal cost of transforming one point cloud into another, capturing structural differences. 
Lastly, Fréchet Inception Distance (FID)~\cite{heusel2017gans} compares the feature distributions of the rendered images from generated 3D shapes with those of actual 3D shape data using a pre-trained Inception network. 
Detailed information and equations for the three evaluation metrics are provided in the supplementary material.

\begin{table}[t]
\begin{center}
\resizebox{\linewidth}{!}{%
\begin{tabular}{@{}lcccccc@{}} \toprule
& \multicolumn{3}{c}{AmateurSketch-3D} & \multicolumn{3}{c}{ProSketch-3D} \\ 
\cmidrule(lr){2-4} \cmidrule(lr){5-7}
    Methods & CD $\downarrow$ & EMD $\downarrow$ & FID $\downarrow$ & CD $\downarrow$ & EMD $\downarrow$ & FID $\downarrow$ \\ \midrule
    Sketch2Mesh~\cite{guillard2021sketch2mesh} & 0.257 & 0.211 & 392.2 & 0.228 & 0.171 & 297.8 \\
    LAS-D~\cite{zheng2023locally} & 0.159 &  0.128 & 197.5 & 0.195 & 0.147 & 193.5 \\
    SENS~\cite{binninger2024sens} &  0.121 & 0.096 & 171.3 & 0.116 & 0.076 & 160.5 \\
    DY3D~\cite{bandyopadhyay2024doodle} & 0.109 & 0.091 & - & 0.093 &  0.087 & - \\
    \midrule
    PASTA (Ours) & \textbf{0.090} & \textbf{0.071} & \textbf{143.9} & \textbf{0.055} & \textbf{0.049} & \textbf{112.2} \\
\bottomrule
\end{tabular}}
\vspace{-0.7em}
\caption{
Quantitative comparison of different methods on the AmateurSketch-3D~\cite{qi2021toward} and ProSketch-3D~\cite{zhong2020towards} datasets. 
The performance is evaluated using CD, EMD, and FID, where lower values indicate better performance.
}
\label{tab:main}
\end{center}
\vspace{-2em}
\end{table}

\subsection{Quantitative Analysis}
In this section, we evaluate the performance of PASTA, on the AmateurSketch-3D~\cite{qi2021toward} and ProSketch-3D~\cite{zhong2020towards}.
When evaluating the AmateurSketch-3D and ProSketch-3D dataset, we train our model using two sketch styles~\cite{chan2022learning, vinker2022clipasso}.
We compare our method with several state-of-the-art approaches~\cite{guillard2021sketch2mesh, zheng2023locally, binninger2024sens, bandyopadhyay2024doodle} that take user sketches as input.
As shown in Tab.~\ref{tab:main}, PASTA consistently outperforms all competing methods in both datasets.
Specifically, our method achieves the lowest CD and EMD values, demonstrating superior geometric accuracy in the 3D shape reconstruction task.
Moreover, PASTA also achieves the best FID scores, further confirming its ability to generate highly realistic and accurate 3D shapes.
Additionally, as shown in Tab.~\ref{tab:main_airlamp},
ours performs particularly well in categories such as airplane and lamp, showcasing its versatility across various object types.
Notably, PASTA excels in preserving missing visual cues in oversimplified sketches, demonstrating improvements over existing methods, and highlighting its effectiveness in capturing detailed 3D structures.
\begin{table}[t]
\begin{center}
\resizebox{\linewidth}{!}{%
\begin{tabular}{@{}lcccccc@{}} \toprule
& \multicolumn{3}{c}{Airplane} & \multicolumn{3}{c}{Lamp} \\ 
\cmidrule(lr){2-4} \cmidrule(lr){5-7}
    Methods & CD $\downarrow$ & EMD $\downarrow$ & FID $\downarrow$ & CD $\downarrow$ & EMD $\downarrow$ & FID $\downarrow$ \\ \midrule
    SENS~\cite{binninger2024sens} &  0.240 & 0.162 & 97.2 & 0.253 & 0.194 & 152.8 \\
    PASTA (Ours) & \textbf{0.188} & \textbf{0.113} & \textbf{90.5} & \textbf{0.195} & \textbf{0.147} & \textbf{125.6} \\
\bottomrule
  \end{tabular}}
\vspace{-0.7em}
\caption{
Quantitative comparison of SENS~\cite{binninger2024sens} on the airplane and lamp datasets obtained with CLIPasso~\cite{vinker2022clipasso}.
PASTA demonstrates superior result across all three metrics in both categories.
}
\label{tab:main_airlamp}
\end{center}
\vspace{-2.5em}
\end{table}

\subsection{Qualitative Analysis}
To further evaluate the performance of our model, we present qualitative results across three categories of objects, including chairs, airplanes, and lamps, as well as its ability to edit shapes. Visual comparisons demonstrate that our model excels in capturing fine-grained details and complex structures more accurately than competing methods.

\myparagraph{Chairs.}
We compare PASTA with state-of-the-art methods, including LAS-D~\cite{zheng2023locally}, SENS~\cite{binninger2024sens}, and DY3D~\cite{bandyopadhyay2024doodle}, to demonstrate superior performance in reconstructing diverse chair designs with accurately capture intricate details and global structures. As shown in Fig.~\ref{fig:Qualitative}, PASTA effectively preserves subtle design elements such as armrests, backrests, and leg structures, which are often oversimplified in sketches. Our 3D shape generations exhibit more refined geometries, confirming the robustness of our approach.
\begin{figure}[t]
    \centering
    \includegraphics[width=0.475\textwidth]{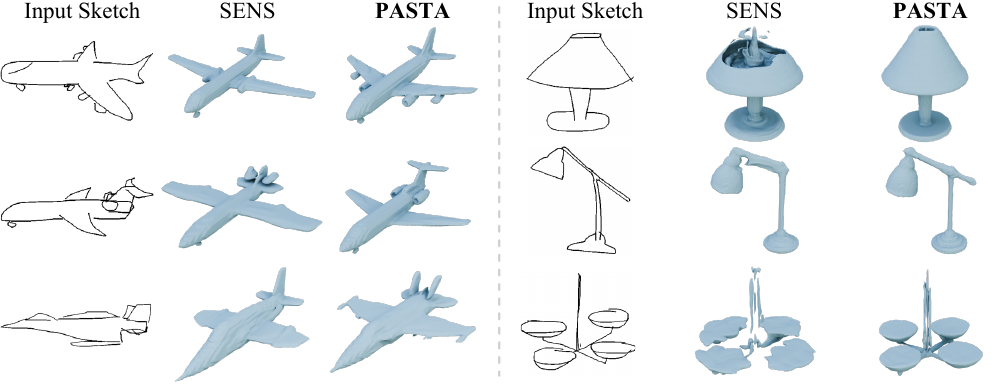}
    \caption{
    3D generation results of SENS~\cite{binninger2024sens} and PASTA for airplane and lamp sketches. 
    The shapes generated by our method accurately capture structural details for both object types.
    }
    \label{fig:airplane}
\end{figure}
\begin{figure}[t]
    \centering
    \includegraphics[width=0.475\textwidth]{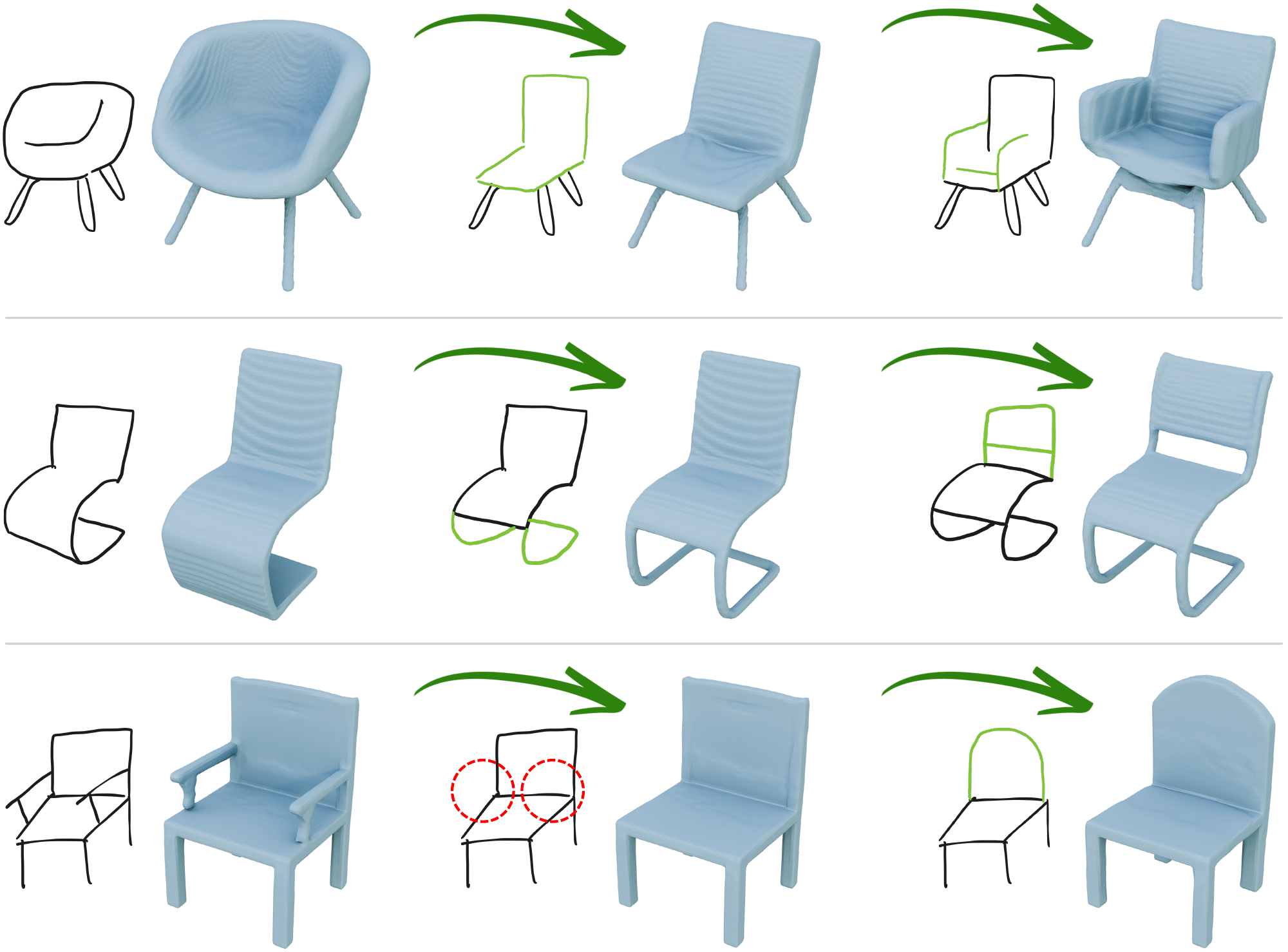}
    \caption{Qualitative results of shape editing. Our model enables intuitive modifications, allowing for controlled adjustments while preserving structural consistency.}
    \label{fig:editing}
    \vspace{-1em}
\end{figure}

\myparagraph{Airplanes.}
For the airplane category, the qualitative results in Fig.~\ref{fig:airplane} show that our method produces more accurate and well-structured reconstructions than the previous model~\cite{binninger2024sens}. PASTA effectively captures essential structural components, including wings, tails, and engines, also maintaining fine-grained details, while SENS often struggles with preserving these details.

\myparagraph{Lamps.}
We also evaluate our model in lamp sketches, where it successfully reconstructs a variety of lamp styles, from simple desk lamps to chandelier-like designs. As depicted in Fig.~\ref{fig:airplane}, PASTA generates plausible 3D shapes that closely align with the input sketches while preserving intricate design elements such as the head shape, lamp base, and delicate frame structures. 

\myparagraph{Shape Editing.}
Our method provides interactive editing capability that enables users to precisely localize and modify the intended parts, thanks to the GMM-based decomposed part-level representation. As illustrated in Fig.~\ref{fig:editing}, PASTA obviously reflects changes in appearance, as well as modifications, additions, and deletions.
For example, when the shapes of the backrest, seat, and legs are altered across all rows, the intended modifications are accurately reflected while maintaining the shape of other parts. Moreover, in the first row, the addition of armrests is clearly depicted, in the second row, the legs and backrest are modified into the desired shapes, and in the third row, their removal is effectively captured.
These experiments demonstrate that our model effectively learns the composition of objects and the relationships of the parts.
\begin{table}[t]
\begin{center}
\resizebox{\linewidth}{!}{%
\begin{tabular}{@{}lcccccc@{}} \toprule
& \multicolumn{3}{c}{AmateurSketch-3D} & \multicolumn{3}{c}{ProSketch-3D} \\ 
\cmidrule(lr){2-4} \cmidrule(lr){5-7}
    Methods & CD $\downarrow$ & EMD $\downarrow$ & FID $\downarrow$ & CD $\downarrow$ & EMD $\downarrow$ & FID $\downarrow$ \\ \midrule
    Baseline & 0.117 & 0.095 & 169.2 & 0.109 & 0.076 & 157.9\\
    w/ IndivGCN & 0.103 & 0.087 & 158.8 & 0.089 & 0.065 & 148.4\\
    w/ PartGCN & 0.109 & 0.090 & 163.1 & 0.093 & 0.071 & 152.0\\
    w/ ISG-Net & \textbf{0.098} & \textbf{0.083} & \textbf{155.7} & \textbf{0.068} & \textbf{0.054} & \textbf{129.2}\\
\bottomrule
\end{tabular}}
\vspace{-0.7em}
\caption{
Ablation study to examine the impact of different GCN configurations, including ``w/ IndivGCN'', ``w/ PartGCN'', and ``w/ ISG-Net'', which refers to the model with both IndivGCN and PartGCN applied.
}
\label{tab:ablation_gsnet}
\end{center}
\vspace{-1.em}
\end{table}

\begin{table}[t]
\begin{center}
\resizebox{\linewidth}{!}{%
\begin{tabular}{@{}lcccccc@{}} \toprule
& \multicolumn{3}{c}{AmateurSketch-3D} & \multicolumn{3}{c}{ProSketch-3D} \\ 
\cmidrule(lr){2-4} \cmidrule(lr){5-7}
    Methods & CD $\downarrow$ & EMD $\downarrow$ & FID $\downarrow$ & CD $\downarrow$ & EMD $\downarrow$ & FID $\downarrow$ \\ \midrule
    Baseline & 0.117 & 0.095 & 169.2 & 0.109 & 0.076 & 157.9\\
    w/ Text & 0.101 & 0.086 & 157.6 & 0.081 & 0.063 & 135.5\\
    w/ ISG-Net & 0.098 & 0.083 & 155.7 & 0.068 & 0.054 & 129.2\\
    PASTA(Ours) & \textbf{0.090} & \textbf{0.071} & \textbf{143.9} & \textbf{0.055} & \textbf{0.049} & \textbf{112.2}\\
\bottomrule
\end{tabular}}
\vspace{-0.7em}
\caption{
Evaluating the impact of text embedding and ISG-Net on model performance. The ``w/ Text'' and ``w/ ISG-Net'' rows show the effects of adding each component individually, while the final PASTA model combines both for improved performance.
}
\label{tab:ablation_contri}
\end{center}
\vspace{-2.em}
\end{table}
\subsection{Ablation Study}
In this study, we conducted two ablation studies to assess the impact of key contributions in our framework by comparing them against a ``Baseline" without any of these components.
In particular, we examine the effectiveness of IndivGCN and PartGCN in capturing structural information from input sketches. 
Additionally, we evaluate the contributions of text embedding and ISG-Net to the overall performance, analyzing how they enhance the quality of the generated 3D shapes.

\myparagraph{Impact of IndivGCN and PartGCN.}
To assess the effectiveness of two GCNs, we compare the following settings in Tab.~\ref{tab:ablation_gsnet}: (i) Baseline, (ii) IndivGCN applied only, (iii) PartGCN applied only, and (iv) ISG-Net (IndivGCN and PartGCN applied together). 
This result shows that using IndivGCN enhances performance by capturing detailed information from individual queries, while the application of PartGCN improves global structure by leveraging part awareness. 
Combining both modules results in even better performance, highlighting their complementary roles in feature extraction.

\myparagraph{Effect of Text-Aligned Prior and ISG-Net.}
We examine the influence of the text-aligned prior and ISG-Net by comparing Tab.~\ref{tab:ablation_contri}: (i) Baseline, (ii) text embedding applied only, (iii) ISG-Net applied only, and (iv) PASTA (text embedding and ISG-Net applied together). 
When applying only the text-aligned prior, a significant performance improvement is observed over the baseline across all three metrics, highlighting the importance of the text condition. Additionally, using only ISG-Net also shows improved performance, which validates the effectiveness of learning structural relationships through the two GCN modules. Ultimately, combining both methods achieves the highest performance, demonstrating that their integration leads to optimal results.
\begin{figure}[t]
    \centering
    \includegraphics[width=0.475\textwidth]{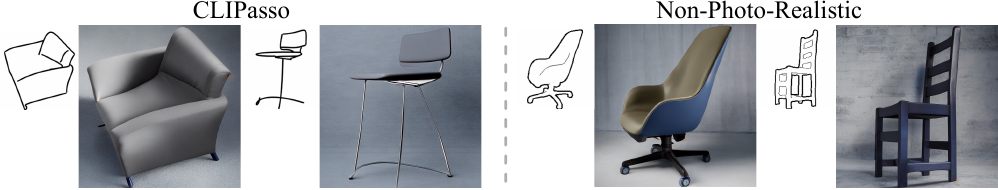}
    \caption{Examples of sketches from CLIPasso~\cite{vinker2022clipasso}, and non-photo-realistic rendering~\cite{chan2022learning}, along with their converted versions using ControlNet~\cite{zhang2023adding}.}
    \label{fig:sketch_sample}
    \vspace{-1.em}
\end{figure}
\begin{figure}[t]
    \centering
    \includegraphics[width=0.475\textwidth]{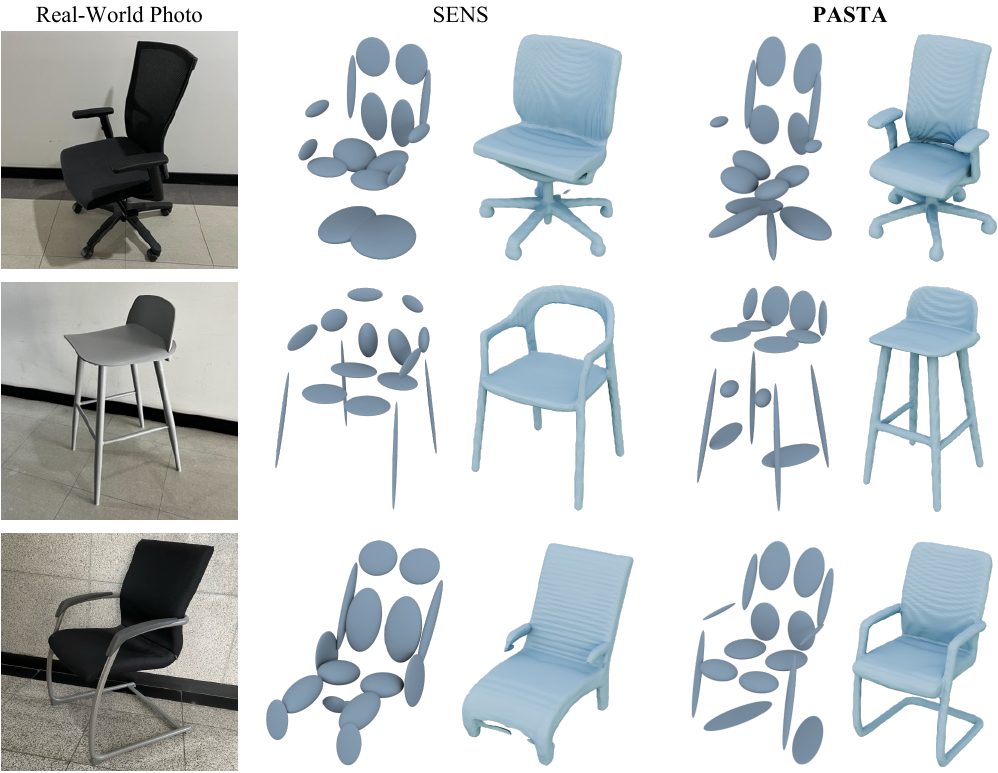}
    \caption{Comparison of outputs when using real-world photos as input. We compare our model with SENS~\cite{binninger2024sens}, showing the improved accuracy and detail in the output. Results demonstrate the ability of PASTA to extend the domain over the sketch.}
    \label{fig:real_image}
    \vspace{-1.em}
\end{figure}
\subsection{Qualitative Results on Real-Image}
To assess the general robustness of the proposed method, we extend qualitative comparisons to real-world image domains. 
Specifically, ControlNet~\cite{zhang2023adding} is utilized to convert grayscale sketches from training datasets~\cite{vinker2022clipasso, chan2022learning} into RGB images while generating colorized outputs that preserve the original sketch's shapes and enhance realism, as illustrated in Fig.~\ref{fig:sketch_sample}. 
To ensure a fair comparison, PASTA is trained on these RGB images alongside SENS~\cite{binninger2024sens}.
During inference, as shown in Fig.~\ref{fig:real_image}, real-world photographs with varying shapes in different places are utilized to further evaluate the model's generalization ability.
The results demonstrate that PASTA achieves superior accuracy and detail in 3D shape generation from real images compared to previous method. 
These findings underscore the effectiveness of the proposed approach in structural analysis and interpretation, validating its applicability beyond grayscale sketch data to RGB real-world images.

\subsection{Effect of Description Types in VLM}
In this section, we further evaluate the impact of text embeddings by comparing results across various description types. 
As shown in Tab.~\ref{tab:ablation_text}, using embeddings that describe only the type of parts (e.g., backrest, armrest, legs) results in improved performance, which demonstrates the effectiveness of text embeddings. 
This approach helps the model recognize basic structures and parts. 
However, listing only the parts is not enough to capture detailed features or complex structures.
Conversely, providing excessively verbose explanations yields some improvement but often leads to overstatements or inaccuracies such as hallucinations, which introduce noise that degrades performance.
Therefore, we use embeddings that specify the exact number and appearance of parts in a sentence, as shown in Fig.~\ref{fig:text}. This approach effectively compensates for missing visual cues and achieves optimal performance on both benchmarks~\cite{zhong2020towards, qi2021toward}.
More detailed examples of part type and verbose descriptions are provided in the supplementary material.

\begin{table}[t]
\begin{center}
\resizebox{\linewidth}{!}{%
\begin{tabular}{@{}lcccccc@{}} \toprule
& \multicolumn{3}{c}{AmateurSketch-3D} & \multicolumn{3}{c}{ProSketch-3D} \\ 
\cmidrule(lr){2-4} \cmidrule(lr){5-7}
    Methods & CD $\downarrow$ & EMD $\downarrow$ & FID $\downarrow$ & CD $\downarrow$ & EMD $\downarrow$ & FID $\downarrow$ \\ \midrule
    Baseline & 0.117 & 0.095 & 169.2 & 0.109 & 0.076 & 157.9\\
    w/ Part Type & 0.103 & 0.088 & 159.5 & 0.092 & 0.069 & 142.8\\
    w/ Verbose & 0.105 & 0.088 & 161.8 & 0.085 & 0.066 & \textbf{134.1}\\
    w/ Single Sent. & \textbf{0.101} & \textbf{0.086} & \textbf{157.6} & \textbf{0.081} & \textbf{0.063} & 135.5\\
\bottomrule
\end{tabular}}
\vspace{-0.7em}
\caption{
Evaluating the impact of various text description formats for text condition, including Part Type, Verbose (more extended sentences), and Single Sentences. 
}
\label{tab:ablation_text}
\end{center}
\vspace{-2.em}
\end{table}
\section{Conclusion}
\vspace{-.2em}
\label{sec:conclusion}

We present PASTA as a significant advancement in sketch-based 3D shape generation, addressing key limitations of prior methods~\cite{binninger2024sens, bandyopadhyay2024doodle, zheng2023locally}. 
Leveraging text conditions from VLM~\cite{liu2023visual} enhances the understanding of object composition, while ISG-Net improves fine-grained detail processing and refines structural coherence through IndivGCN and PartGCN.
Additionally, PASTA also enables more precise and flexible part-level editing. 
Moreover, PASTA proves its versatility by enabling 3D shape generation not only from sketches but also from real-world images, emphasizing its scalability.
Comprehensive experiments, including both qualitative and quantitative analyses, PASTA is established as a significant advancement in 3D shape generation, demonstrating its superiority in structural coherence, detail preservation, and part-level flexibility.

\myparagraph{Limitations \& Future Work.}
Our approach uses a shape decoder, SPAGHETTI~\cite{hertz2022spaghetti}, enabling fast training and flexible generation. However, it is limited to a few classes and a fixed number of Gaussians. Future work may explore removing these constraints and incorporating a dedicated implicit decoder to support a wider range of categories.
{
    \small
    \bibliographystyle{ieeenat_fullname}
    \bibliography{ICCV2025-Author-Kit-Feb/main}
    
}

\clearpage
\setcounter{page}{1}
\setcounter{section}{0}
\setcounter{figure}{0} 
\setcounter{table}{0} 
\maketitlesupplementary

\section*{Overview}
The supplementary material provides a detailed description of our proposed method, PASTA, for sketch-to-3D shape generation.
It begins with the experimental setup in Section~\ref{supp:section1}, which outlines the data sets~\cite{vinker2022clipasso, chan2022learning, zhong2020towards, qi2021toward} used for training and evaluation, the details of the implementation, and the evaluation metrics used to assess the quality of the 3D shapes generated.
In Section~\ref{supp:TVTD}, we elaborate details on the Text-Visual Transformer Decoder, which plays a crucial role in integrating text and visual information to enhance the semantic understanding of input hand-drawn sketches. 
In Section~\ref{supp:ISG}, we analyze Integrated Structure-Graph Network (ISG-Net), a graph-based refinement module designed to improve the structural coherence of the generated 3D models. 
Finally, in Section~\ref{supp:section4}, we present more qualitative results, demonstrating the effectiveness of PASTA in generating high-quality 3D shapes that faithfully preserve the intended design elements from sketches. 

\section{Experimental Setup}\label{supp:section1}
\subsection{Dataset} 
\begin{figure}[t]
    \centering
    \vspace{-1.em}
    \includegraphics[width=\linewidth]{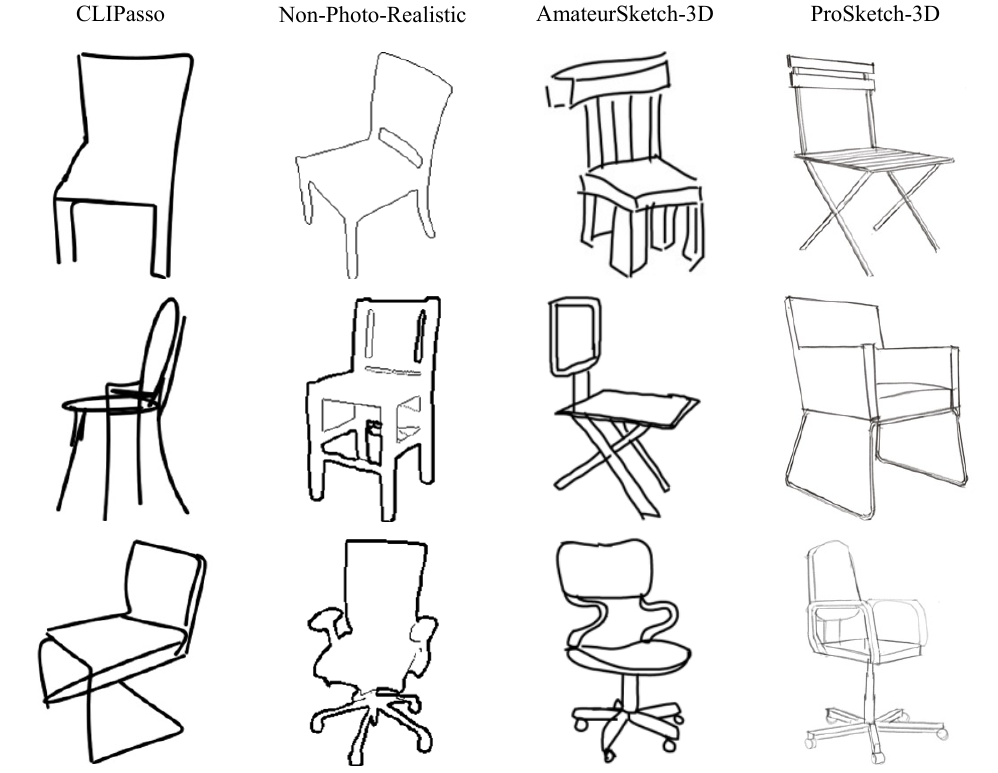}
    \vspace{-1.5em}
    \caption{Sample images from the datasets used in our experiments, including CLIPasso~\cite{vinker2022clipasso}, non-photo-realistic renderings~\cite{chan2022learning}, AmateurSketch-3D~\cite{qi2021toward}, and ProSketch-3D~\cite{zhong2020towards}. These datasets capture a range of sketching styles, from highly abstract representations to detailed, expert-drawn sketches.}
    \label{fig:supp1}
    \vspace{-0.4em}
\end{figure}

To evaluate the effectiveness of our proposed method, we utilize multiple datasets~\cite{vinker2022clipasso, chan2022learning, zhong2020towards, qi2021toward} representing diverse sketching styles and their corresponding 3D shapes. 
Specifically, the datasets used in this work include CLIPasso~\cite{vinker2022clipasso} and non-photo-realistic renderings~\cite{chan2022learning}, both of which provide diverse stylistic representations of objects. CLIPasso generates abstract, highly simplified depictions, while non-photo-realistic renderings produce more realistic sketches compared to CLIPasso. 
For evaluation, we used the AmateurSketch-3D~\cite{qi2021toward} and ProSketch-3D~\cite{zhong2020towards} datasets. AmateurSketch-3D consists of freehand sketches drawn by non-experts, often exhibiting variability in proportions and details, whereas ProSketch-3D comprises highly detailed, expert-drawn sketches that adhere closely to object structures. Each dataset contributes a distinct perspective on object depiction, spanning a spectrum from abstract simplifications to highly refined artistic renderings. Fig.~\ref{fig:supp1} presents sample sketches from these datasets~\cite{qi2021toward, zhong2020towards, chan2022learning, vinker2022clipasso}, highlighting differences in abstraction level, detail, and style of chair sketches.

\begin{figure}[t]
    \centering
    \vspace{-1.em}
    \includegraphics[width=\linewidth]{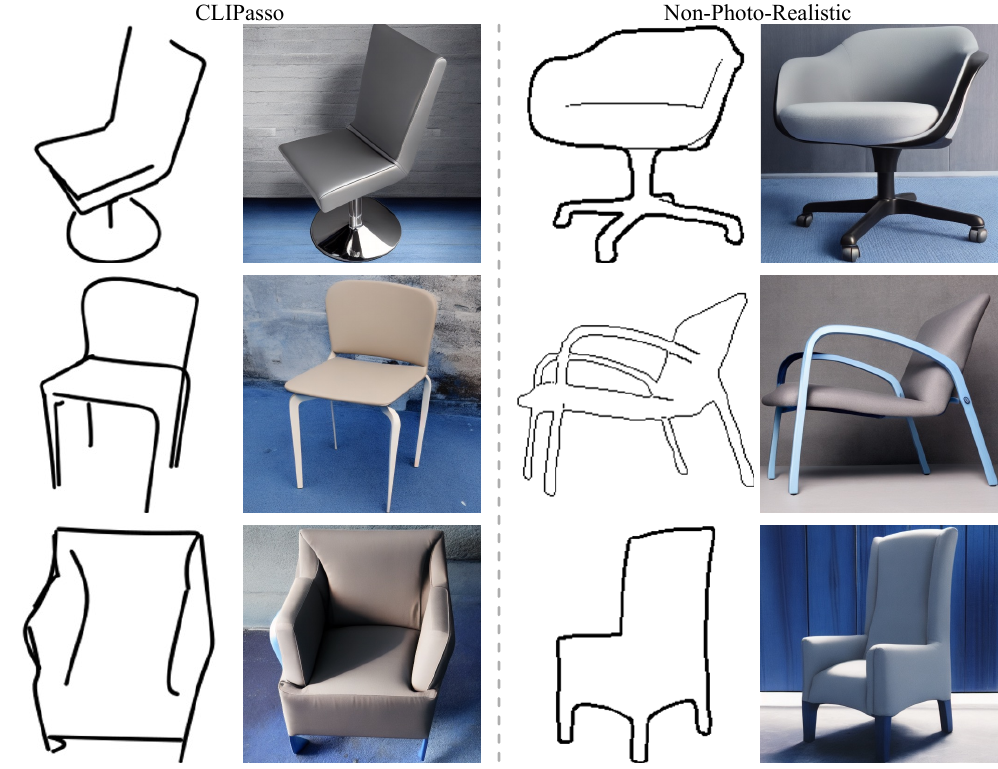}
    \vspace{-1.5em}
    \caption{Illustration of sketches processed using ControlNet~\cite{zhang2023adding}. The transformed sketches retain the structural essence of the originals while incorporating enhanced realism.}
    \label{fig:supp2}
    \vspace{-1.em}
\end{figure}

In addition to these datasets~\cite{vinker2022clipasso, chan2022learning}, we convert realistic images using ControlNet~\cite{zhang2023adding} to further analyze our approach on real-world photo based 3D shape generation. Specifically, we apply ControlNet to sketches from CLIPasso and non-photo-realistic renderings, producing enhanced versions that preserve the original structural essence of the sketches while incorporating greater realism. Using ControlNet to guide image generation with fine-grained control, we ensure that the generated results retain the fundamental characteristics of the input sketches while resembling realistic RGB images. This allows us to bridge the gap between abstract grayscale representations and photorealistic RGB images, offering a richer set of inputs for 3D generation. 
The generated results are displayed in Fig.~\ref{fig:supp2}, demonstrating how ControlNet enhances the visual fidelity of sketches while maintaining their structural shape through examples in the realistic RGB images.
\subsection{Implementation Details}
\myparagraph{Training Configuration.}
For the training and evaluation of our model, we adopt a carefully designed experimental setup to ensure robust performance and fair comparisons with existing methods. 
Our model is trained on a single RTX 3090 GPU for approximately 38 hours for the chair category.
The training process follows a batch size of 16 and employs an initial learning rate of ${10^{-4}}$, which is dynamically adjusted using the OneCycle learning rate scheduler~\cite{smith2019super} to facilitate stable convergence. 
This scheduler gradually increases the learning rate in the early stages of training before decaying it towards the end, preventing premature convergence and improving generalization. 
The model is optimized using the Adam optimizer, and we train for 650 epochs, ensuring sufficient iterations for convergence while mitigating the risk of overfitting.
\clearpage
\myparagraph{Architectural Specifications.}
In addition to these experimental settings, we provide detailed specifications regarding the architectural components of PASTA. We set the number of queries and Gaussian mixture models to $N = 16$. Within the PartGCN module, the number of graph clustering groups is set to $K = 4$, ensuring effective part-wise feature aggregation. The overall model architecture is designed with two layers in the graph-convolution network (GCN), allowing for a balance between expressiveness and computational efficiency. As described in Equation (9) of the main paper, the weighting factor $\alpha$ in IndivGCN and PartGCN is set to 0.8 based on experimental results. 
Furthermore, the loss weights for different components, including $\lambda_{\text{align}}$, $\lambda_{\text{indiv}}$, and $\lambda_{\text{part}}$, are set to 1.0, 0.1, and 0.1, respectively, following the formulation in Equation (11) of the main paper, ensuring optimal performance in various sketch styles and object categories.

\subsection{Evaluation Metrics}
For evaluation, we employ three metrics to assess the quality of the 3D shapes generated. Chamfer distance (CD) quantifies the accuracy of the point-wise reconstruction by measuring the discrepancy between the predicted point clouds set $P$ and the ground-truth point clouds set $G$: 
\setlength{\abovedisplayskip}{0.5em}  
\setlength{\belowdisplayskip}{0.5em}  
\begin{equation}
\text{CD}(P, G) = \frac{1}{|P|} \sum_{p\in P} \min_{g\in G} \|p - g\|^2_2 + \frac{1}{|G|} \sum_{g \in G} \min_{p \in P} \|p - g\|^2_2.
\end{equation}
The Earth Mover’s Distance (EMD) quantifies structural differences by determining the minimum cost required to transform one point cloud into another, based on an optimal correspondence $\pi \in \Pi(P, G)$. Here, $\Pi(P, G) \in \mathbb{R}^{n \times m}$ consists of elements in the range between 0 and 1, such that the sum of each row and each column equals one.
\begin{equation} 
\text{EMD}(P, G) = \min_{\pi \in \Pi(P,G)} \sum_{i=1}^{n} \sum_{j=1}^{m} \pi_{i,j} \|p_i - g_j\|. 
\end{equation} 
For the calculation of CD and EMD, 2,048 points are sampled from both the ground truth mesh and the generated mesh.
\begin{figure}[t]
    \centering
    \vspace{-1.em}
    \includegraphics[width=\linewidth]{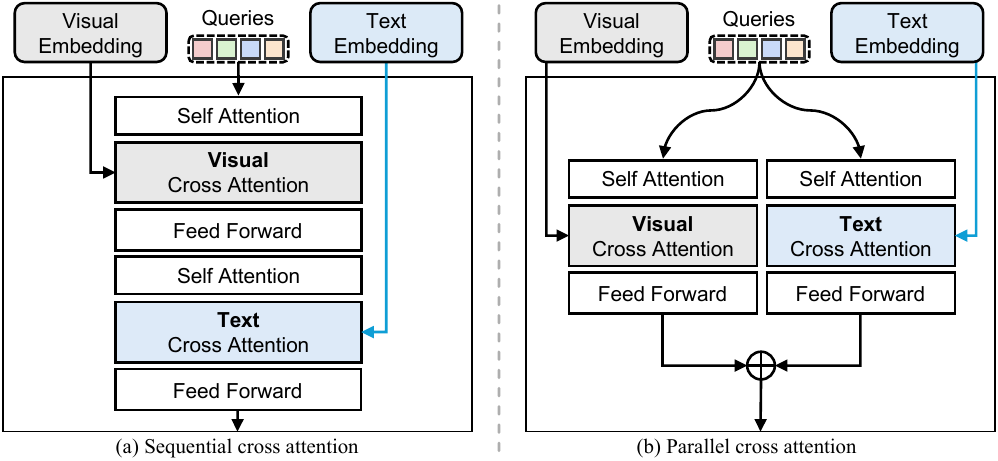}
    \vspace{-1.5em}
    \caption{
    Comparison of two architectures for the Text-Visual Transformer Decoder: (a) sequential cross-attention and (b) parallel cross-attention mechanisms.
    }
    \label{fig:supp3}
\end{figure}
\begin{table}[t]
\begin{center}
\resizebox{\linewidth}{!}{%
\begin{tabular}{@{}lcccccc@{}} \toprule
& \multicolumn{3}{c}{AmateurSketch-3D} & \multicolumn{3}{c}{ProSketch-3D} \\ 
\cmidrule(lr){2-4} \cmidrule(lr){5-7}
    Methods & CD $\downarrow$ & EMD $\downarrow$ & FID $\downarrow$ & CD $\downarrow$ & EMD $\downarrow$ & FID $\downarrow$ \\ \midrule
    (a) Sequential & \textbf{0.090} & \textbf{0.071} & \textbf{143.9} & \textbf{0.055} & \textbf{0.049} & \textbf{112.2} \\
    (b) Parallel &  0.095 & 0.074 & 150.2 & 0.071 & 0.061 & 120.7 \\ 
\bottomrule
\end{tabular}}
\vspace{-0.7em}
\caption{
Quantitative comparison of the two Text-Visual Transformer Decoder architectures. (a) Sequential method consistently outperforms  (b) parallel method across multiple evaluation metrics, highlighting the benefits of first enriching visual embeddings before merging them with textual information.
}
\label{tab:supp_ca2}
\end{center}
\vspace{-2.em}
\end{table}

Finally, the Fréchet Inception Distance (FID)~\cite{heusel2017gans} is used to assess the realism of the generated shapes by comparing the feature distributions of the rendered 3D models with those of real-world reference data:
\begin{equation} 
\text{FID} = \frac{1}{20} \sum_{i=1}^{20} \left( \|\mu_i - \mu_i^{\prime}\|^2_2 + \text{Tr} \left( \Sigma_i + \Sigma_i^{\prime} - 2\sqrt{\Sigma_i \Sigma_i^{\prime}} \right) \right).
\end{equation}
To compute the FID score, we first sample 20 different views and render both the ground truth shape $S$ the generated shape $S^{\prime}$. The features are then extracted from these images using the Inception-V3 network~\cite{szegedy2016rethinking}, 
which maps each image to a probability distribution across 1,000 classes. From this distribution, we compute the mean $\mu_i$ and the covariance matrix $\Sigma_i$ for each image $i$. These statistics are then used to calculate the final FID score. To improve the accuracy of the calculation, we employ the shading-image-based FID metric.

By systematically integrating these experimental configurations, our approach is designed to generate high-fidelity 3D shape generations while maintaining structural consistency across different types of input sketches. The following sections further analyze our results, highlighting details in qualitative and quantitative improvements. 

\section{Details of Text-Visual Transformer Decoder} \label{supp:TVTD}
The Text-Visual Transformer Decoder is a key contribution in our framework, responsible for integrating text priors with visual features to enhance 3D shape generation from sketches. In this section, we present the architectural design of the decoder, analyze the impact of different vision-language models (VLMs)~\cite{radford2021learning, li2023blip, liu2023visual}, and evaluate the role of input text descriptions in guiding the generation process.

\begin{figure}[t]
    \centering
    \vspace{-1.em}
    \includegraphics[width=\linewidth]{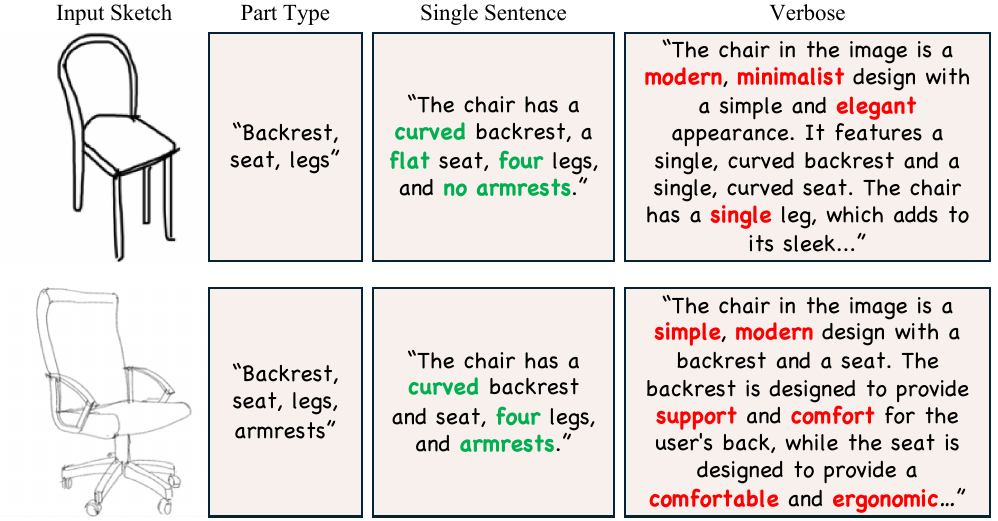}
    \vspace{-1.5em}
    \caption{
    Examples of different text description styles used in the VLM~\cite{liu2023visual}. The three types include Part Type, Single Sentence, and Verbose descriptions.
    }
    \label{fig:supp4}
    \vspace{-0.5em}
\end{figure}
\begin{table}[t]
\begin{center}
\resizebox{\linewidth}{!}{%
\begin{tabular}{@{}lcccccc@{}} \toprule
& \multicolumn{3}{c}{AmateurSketch-3D} & \multicolumn{3}{c}{ProSketch-3D} \\ 
\cmidrule(lr){2-4} \cmidrule(lr){5-7}
    Methods & CD $\downarrow$ & EMD $\downarrow$ & FID $\downarrow$ & CD $\downarrow$ & EMD $\downarrow$ & FID $\downarrow$ \\ \midrule
    CLIP~\cite{radford2021learning} &  0.101 & 0.080 & 152.1 & 0.066 & 0.053 & 117.7 \\
    BLIP2 T5~\cite{li2023blip} &  0.095 & 0.073 & 147.4 & 0.061 & 0.051 & 115.6 \\
    LLaVa-13B~\cite{liu2023visual} &  0.091 & \textbf{0.071} & 144.3 & \textbf{0.054} & 0.050 & 113.9 \\
    \midrule
    LLaVa-7B~\cite{liu2023visual} & \textbf{0.090} & \textbf{0.071} & \textbf{143.9} & 0.055 & \textbf{0.049} & \textbf{112.2} \\
\bottomrule
\end{tabular}}
\vspace{-0.7em}
\caption{
Performance comparison of different VLMs~\cite{liu2023visual, li2023blip, radford2021learning} in our framework. LLaVa-7B~\cite{liu2023visual} achieves the best balance between CD, EMD, and FID that making it the most suitable choice for enhancing text-guided 3D shape generation.
}
\label{tab:supp_vlms}
\end{center}
\vspace{-2em}
\end{table}
\subsection{Architectural Design of Text-Visual Transformer Decoder}
Fig.~\ref{fig:supp3} presents two different architectures to integrate text and visual condition within our Text-Visual Transformer Decoder. In method (a), the visual embedding is first processed through a cross-attention mechanism before being fused with text embeddings via a second cross-attention operation. In contrast, method (b) applies cross-attention separately to both visual and text embeddings before merging them later in the process. Tab.~\ref{tab:supp_ca2} compares the performance of these two approaches, showing that method (a) consistently outperforms method (b). 
The key advantage of method (a) is that the initial cross-attention incorporates visual condition into the query, allowing the subsequent text cross-attention to extract more relevant information. Since the text embeddings extracted from the VLM~\cite{liu2023visual} are aligned with the visual latent space, integrating the sketch embedding into the query first aids in the consolidation of corresponding part-specific information. In contrast, as in method (b), applying cross-attention between a query devoid of visual cues and the text embeddings may lead to unintended information exchange.
This results in improved alignment between sketches and their corresponding 3D shapes, leading us to adopt method (a) in our framework.

\subsection{Impact of Vision-Language Models}
To assess the influence of different VLMs~\cite{radford2021learning, li2023blip, liu2023visual} on our framework, we conduct a comparative analysis using various VLMs, as summarized in Tab.~\ref{tab:supp_vlms}.
When employing CLIP~\cite{sanghi2023clip}, only marginal performance improvements are observed, likely due to its categorical training and alignment, which are specifically optimized for classification tasks. In contrast, models that excel in image captioning and visual question answering (VQA) yield more substantial enhancements. Notably, BLIP2~\cite{li2023blip}, lacking explicit fine-tuning on instruction data, frequently produces suboptimal outcomes compared to LLaVA-7B/13B. For instance, it sometimes generates responses that are irrelevant to the user’s instructions (e.g., \textit{“a chair is a piece of furniture with a seat and backrest.”}) or omits critical components, such as the number of legs and the presence of armrests (e.g., \textit{“a chair with a backrest and seat that are shaped in the form of a horseshoe.”}). 
In contrast, the LLaVA-7B model consistently provides responses that comprehensively incorporate the necessary elements. While LLaVA-13B delivers descriptions similar to those of the 7B model, its performance is marginally inferior. Considering model size, computational cost, and overall performance balance, we therefore adopt the 7B model in this work.

\vspace{0.4em}
\myparagraph{Effect of Description Types on Text Embeddings.}
The quality and structure of text descriptions play a valuable role in guiding the 3D generation process. 
Fig.~\ref{fig:supp4} provides examples of different description styles, corresponding to part type based descriptions, single sentence descriptions, and verbose descriptions. 
For instance, in the ~\textbf{Verbose}, the description of such discrepancies arises due to overemphasis on stylistic attributes like ``modern", ``minimalist", and ``elegant" which can mislead the 3D generation model. 
On the other hand, ~\textbf{Part Type}, such as ``Backrest, seat, legs, armrests," provides a fundamental structural understanding. 
However, they lack details about the shape and number of components, which are crucial for precise shape generation.
In contrast, ~\textbf{Single Sentence} strikes a balance between clarity and informativeness. 
They specify the number and form of object components while avoiding unnecessary complexity.
Our analysis, supported by Table 5 in the main paper, reveals that verbose descriptions often introduce hallucinations and excessive modifiers leading to inconsistencies in shape generation. 
As a result, our framework adopts single sentence descriptions to maximize the accuracy and reliability of text-guided 3D shape generation, by maintaining specificity while avoiding hallucinations.
\clearpage
\begin{figure}[t]
    \centering
    \vspace{-1.em}
    \includegraphics[width=\linewidth]{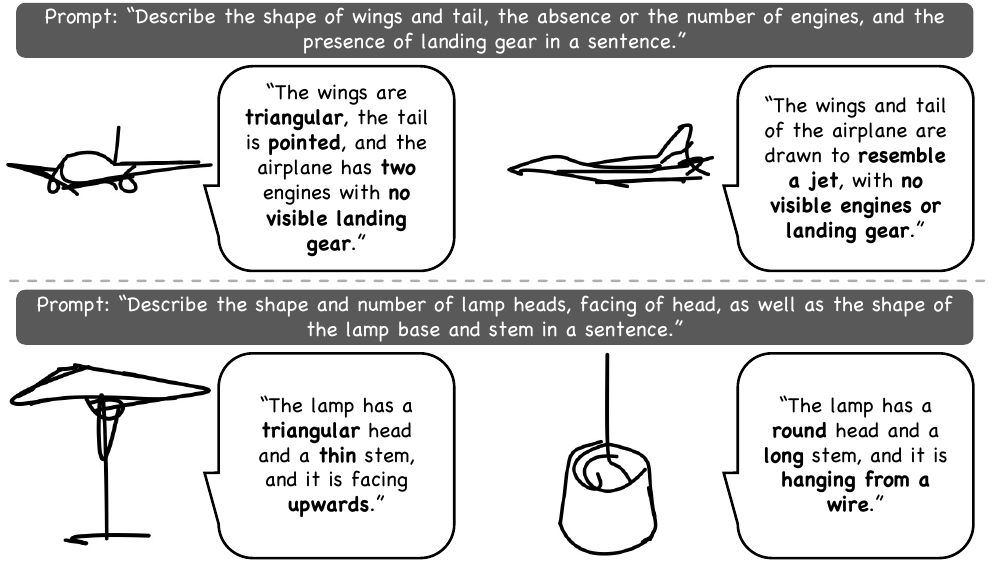}
    \vspace{-1.5em}
    \caption{
    This figure presents prompts used for airplane and lamp sketches, along with the text descriptions produced by the VLM~\cite{liu2023visual}. The generated descriptions capture the key structural features of each sketch, providing valuable semantic information.
    }
    \label{fig:supp5}
    \vspace{-1.2em}
\end{figure}
\myparagraph{Text-Based Descriptions and Generated Outputs.}
We extend our analysis beyond the chair category to include airplane and lamp, as shown in Fig.~\ref{fig:supp5}. 
The results demonstrate that VLM~\cite{liu2023visual} effectively captures fine-grained structural details across different object types, showcasing its ability to perform effectively across diverse shape categories. 
By utilizing text-aligned priors, PASTA successfully distinguishes between distinct design elements, such as the wing and fuselage in airplanes or the lamp head and base in lamps. These findings highlight the robustness of text-visual integration strategy, confirming its effectiveness in enhancing the semantic understanding of sketches across multiple categories.
\vspace{-0.5em}
\section{Analysis of ISG-Net} \label{supp:ISG}
The Integrated Structure-Graph Network (ISG-Net) is designed to refine the structural consistency of 3D shape generation by incorporating graph-based reasoning. The network consists of two key modules: IndivGCN, which focuses on fine-grained feature extraction, and PartGCN, which aggregates and refines part-level information. 
In this section, we first explore related work on Graph Neural Networks (GNNs) and then analyze the effects of the clustering method used in PartGCN, the impact of the number of clusters $K$, and the balance between IndivGCN and PartGCN using the parameter $\alpha$.

\begin{figure}[t]
    \centering
    \vspace{-1.em}
    \includegraphics[width=\linewidth]{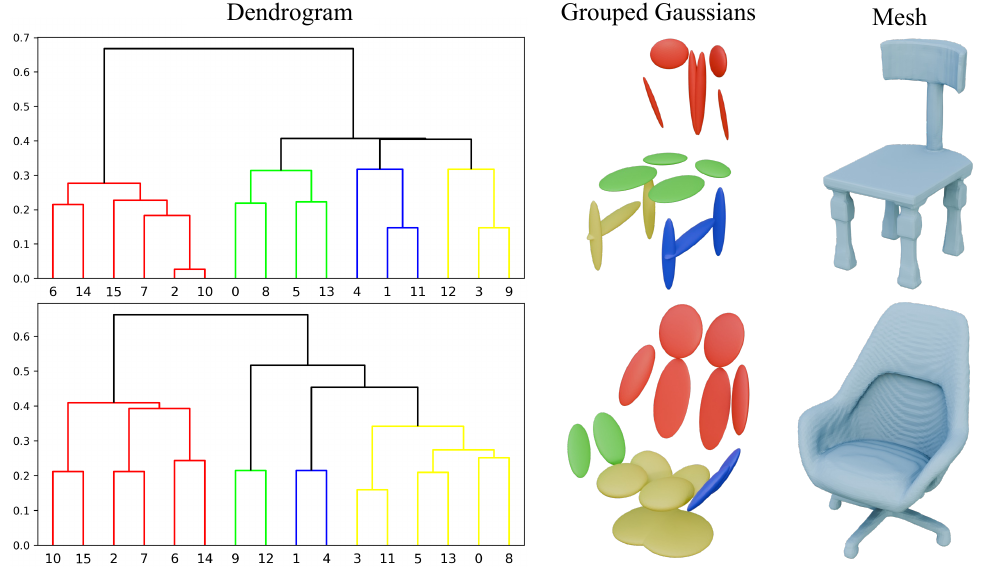}
    \vspace{-1.5em}
    \caption{
    Dendrogram~\cite{zhao2005hierarchical} and visualization of Gaussians with the mesh illustrating the clustering process in PartGCN, where $N = 16$ nodes are grouped into $K=4$ clusters to enhance part-wise feature aggregation.
    }
    \label{fig:supp6}
    \vspace{-1em}
\end{figure}
\subsection{Releated Work on Graph Neural Networks}
Graph neural networks (GNNs)~\cite{scarselli2008graph} model complex structured data as nodes and edges, thus effectively learning relationships within the data. Early work focused on message-passing mechanisms to aggregate information among nodes, and later convolutional neural networks (CNNs) were generalized to graph-structured data, leading to the development of graph convolutional networks (GCNs) that facilitate information exchange between neighboring nodes. GCNs can be classified into spatial-based methods~\cite{niepert2016learning, bruna2013spectral}, which apply trainable filters directly to connected nodes, and spectral-based methods~\cite{kipf2016semi, henaff2015deep}, which define locality through spectral analysis. Such graph-based neural networks have been applied not only to social networks~\cite{hamilton2017inductive} but also to various computer vision tasks~\cite{xu2017scene, landrieu2018large, wang2019learning, teney2017graph, johnson2018image, bae2022learning, kim2024enhanced}. In this work, we apply GCNs to Gaussian mixture models (GMMs) for 3D shape representation. We define the graph structure by producing an adjacency matrix based on the distances between Gaussians in 3D space, thereby enabling information exchange between adjacent nodes for more structurally accurate shape generation and reconstruction. Furthermore, by grouping similar nodes and performing additional graph operations, we achieve robust structural learning.

\subsection{Clustering Method for PartGCN}
To perform PartGCN operations, $N$ queries and Gaussians must be assigned to $K$ subsets. To achieve this, we employ a hierarchical clustering technique~\cite{zhao2005hierarchical} to partition the parts. 
This clustering is applied to both the pseudo-ground truth and the predicted queries. 
For pseudo-ground truth, clustering is performed based on $\mathbf{A}_I \in \mathbb{R}^{N \times N}$ to form $K$ groups, and the coordinates of the Gaussians within each group are aggregated using average pooling to obtain a representative coordinate. 
The distances between these coordinates are then computed to produce the final pseudo-ground truth adjacency matrix $\mathbf{A}_P \in \mathbb{R}^{K \times K}$. 
Simultaneously, predicted queries are pooled for each group to obtain $\mathbf{Q}_{P} \in \mathbb{R}^{K \times d}$, which is processed through $f_{\text{part}}$ and subsequently subjected to an dot product operation to produce the predicted adjacency matrix $\mathbf{\tilde{A}}_P \in \mathbb{R}^{K \times K}$.
As illustrated in Fig.~\ref{fig:supp6}, the clustering process is depicted through a dendrogram~\cite{zhao2005hierarchical}, where $N$ nodes are grouped into $K$ clusters. 
In particular, it flexibly groups dynamically varying Gaussians (\eg, those located near the legs versus those near the armrests). 
This method enables PartGCN to accurately identify and select relevant structural parts, ensuring that the network focuses efficiently on the details of the part level while maintaining overall structural integrity.
\clearpage
\subsection{Effect of Cluster Number $K$}
The experimental results on varying the number of clusters $K$ reveal important insights into the impact of cluster selection on the performance of our model. When $K$ is too small, there is insufficient granularity in the clustering process, causing PartGCN to lose critical structural relationships. 
As a result, the overall structural integrity of the output suffers and the model does not adequately represent the coherence of the object.
On the other hand, when $K$ is excessively large, the model faces a different challenge. 
Although more clusters may provide a higher resolution of detail, it also introduces redundancy and excessive complexity. 
In this case, PartGCN becomes overloaded with unnecessary information, leading to a loss of synergy with IndivGCN. 
This imbalance occurs because excessive detail hampers the effective integration of local features, causing IndivGCN and PartGCN to work against each other rather than complement each other. The model's performance, therefore, suffers from a lack of coherence between the fine-grained details and the global structure.
As shown in Tab.~\ref{tab:supp_k}, $K$=4 ensures that both components work harmoniously, leading to the generation of high-quality 3D shapes without sacrificing performance or structural integrity.
This process allows PartGCN to contribute effectively to the overall structure while maintaining the balance with IndivGCN.

\begin{table}[t]
\begin{center}
\resizebox{0.6\linewidth}{!}{%
\begin{tabular}{@{}lccc@{}} \toprule
& \multicolumn{3}{c}{AmateurSketch-3D} \\
\cmidrule(lr){2-4} 
    Methods & CD $\downarrow$ & EMD $\downarrow$ & FID $\downarrow$ \\ \midrule
    K = 2 &  0.098 & 0.081 & 160.2  \\
    K = 4 & \textbf{0.090} & \textbf{0.071} & \textbf{143.9} \\
    K = 6 &  0.093 & 0.076 & 149.0  \\
    K = 8 &  0.094 & 0.078 & 156.4  \\
\bottomrule
\end{tabular}}
\vspace{-0.7em}
\caption{
Performance comparison for different cluster numbers $K$. Too few clusters weaken structural representation, while too many reduce synergy with IndivGCN.
}
\label{tab:supp_k}
\end{center}
\vspace{-2em}
\end{table}

\subsection{Influence of $\alpha$ in IndivGCN and PartGCN}
In our experiments, we use the parameter $\alpha$ as a weighting coefficient for the outputs of IndivGCN and PartGCN, as described in Equation (9) of the main paper. 
Specifically, $\alpha$ scales the output of IndivGCN, which captures fine-grained local features, while the complement $(1 - \alpha)$ scales the output of PartGCN, which encodes global structural elements.
The experimental results, shown in Tab.~\ref{tab:supp_alpha}, demonstrate that the best performance is achieved when $\alpha$ is set to 0.8, striking the optimal synergy between these two components. 
In this configuration, IndivGCN is weighted more heavily, which allows it to preserve fine-grained local details that are crucial for the accurate representation of intricate features.
At the same time, PartGCN is still capable of effectively contributing to the integrity of the global structure by encoding broader spatial relationships and consistency at the part level, as it is scaled by $(1 - \alpha)$.
This balance is key to ISG-Net success, while IndivGCN focuses on refining local geometry and intricate details, PartGCN provides essential structural support, preventing shape distortions, and preserving overall coherence. 
Consequently, this leads to the best overall performance of ISG-Net, enabling it to generate 3D shapes that are rich in fine-grained detail and structurally sound, preserving oversimplified sketches.
\begin{table}[t]
\begin{center}
\resizebox{0.6\linewidth}{!}{%
\begin{tabular}{@{}lccc@{}} \toprule
& \multicolumn{3}{c}{AmateurSketch-3D} \\
\cmidrule(lr){2-4} 
    Methods & CD $\downarrow$ & EMD $\downarrow$ & FID $\downarrow$ \\ \midrule
    $\alpha = 0.0$ & 0.095 & 0.084 & 157.0  \\ 
    $\alpha = 0.2$ & 0.095 & 0.081 & 155.4  \\ 
    $\alpha = 0.4$ &  0.092 & 0.077 & 151.8  \\ 
    $\alpha = 0.6$ &  0.094 & 0.077 & 152.1  \\ 
    $\alpha = 0.8$ & \textbf{0.090} & \textbf{0.071} & \textbf{143.9} \\
    $\alpha = 1.0$ &  0.092 & \textbf{0.071} & 145.3  \\ 
\bottomrule
\end{tabular}}
\vspace{-0.7em}
\caption{
Effect of $\alpha$ on IndivGCN and PartGCN balance. 
The best performance is achieved at $\alpha=0.8$, optimizing detail and structure integration.
}
\label{tab:supp_alpha}
\end{center}
\vspace{-2em}
\end{table}

\section{Qualitative Results}\label{supp:section4}
\begin{figure*}[t]
    \centering
    \vspace{-1.em}
    \includegraphics[width=\linewidth]{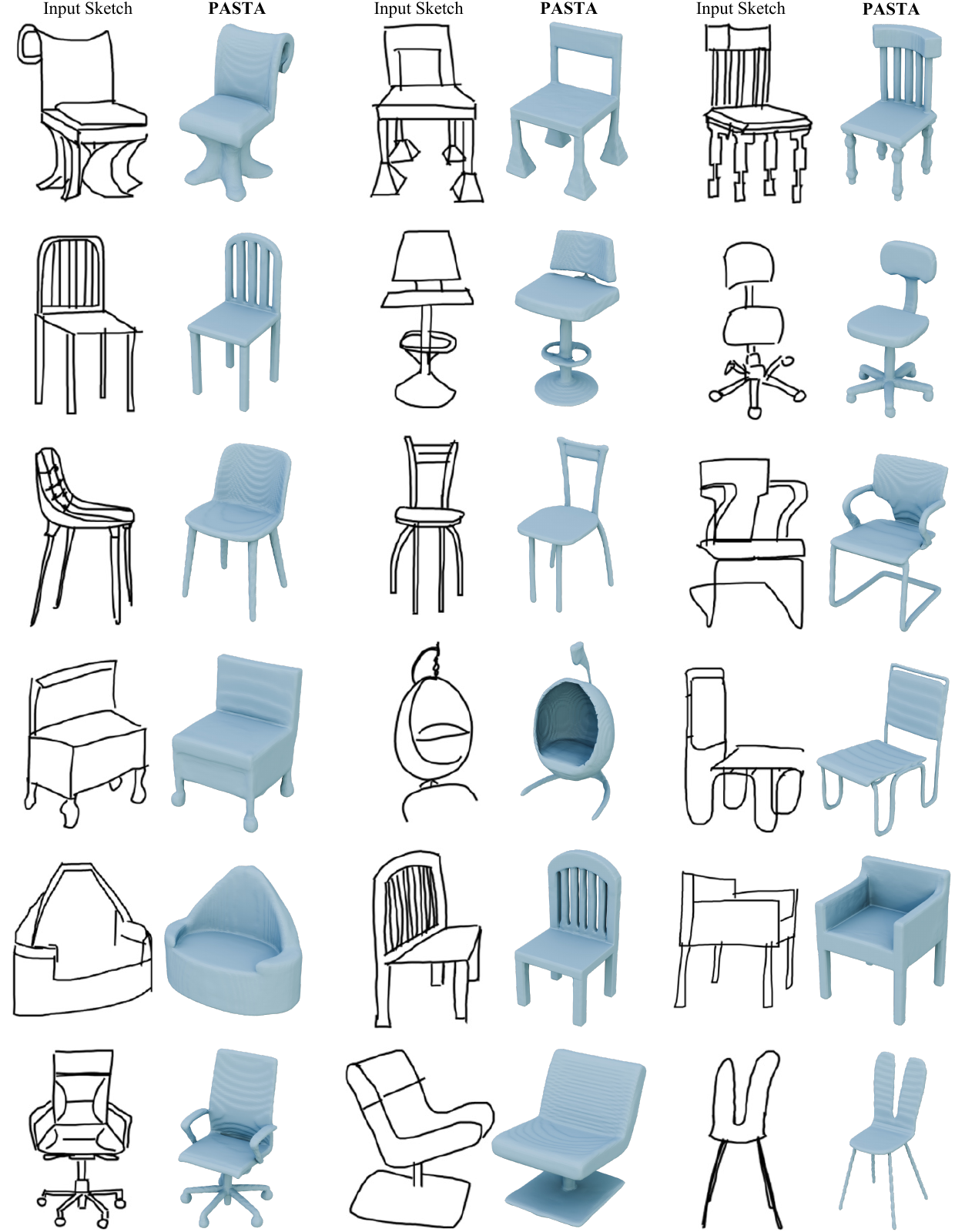}
    \vspace{-1.5em}
    \caption{
    Qualitative results for chair reconstructions, demonstrating the preservation of key structural elements such as leg orientation, backrest curvature, and armrest presence.
    }
    \label{fig:supp7}
    \vspace{-1.em}
\end{figure*}

\begin{figure*}[t]
    \centering
    \vspace{-1.em}
    \includegraphics[width=\linewidth]{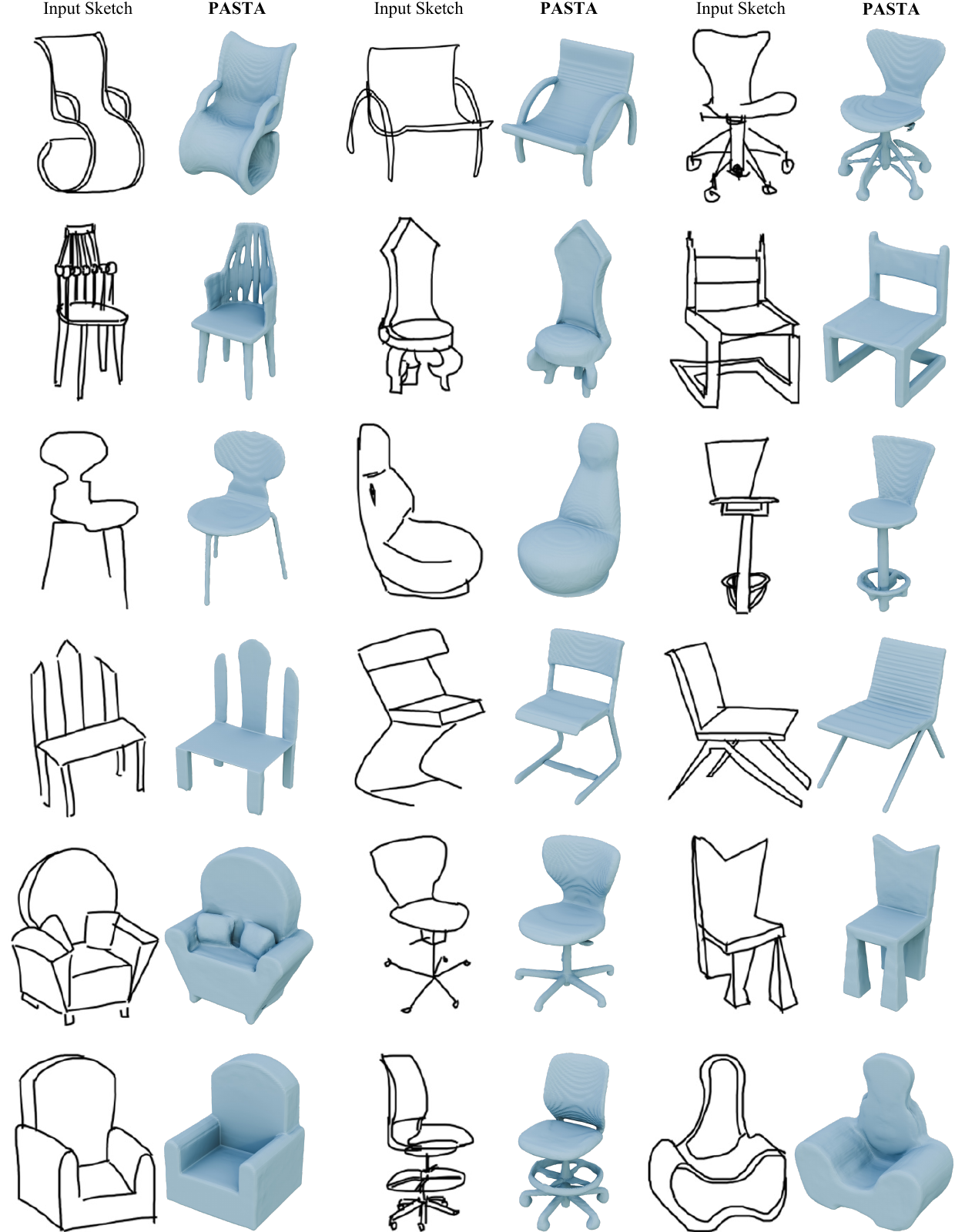}
    \vspace{-1.5em}
    \caption{
    Qualitative results for chair reconstructions, demonstrating the preservation of key structural elements such as leg orientation, backrest curvature, and armrest presence.
    }
    \label{fig:supp8}
    \vspace{-1.em}
\end{figure*}

\begin{figure*}[t]
    \centering
    \vspace{-1.em}
    \includegraphics[width=\linewidth]{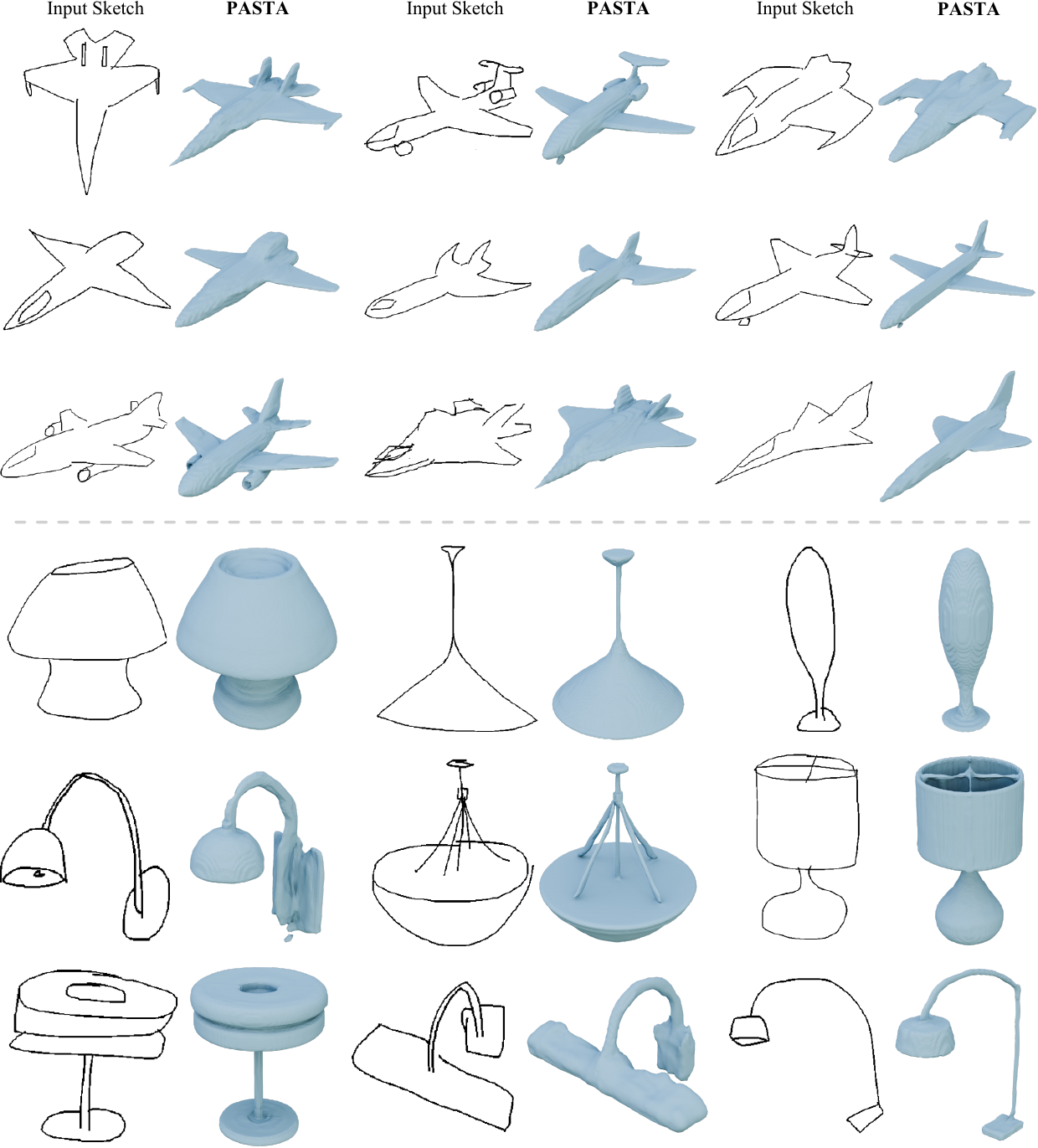}
    \vspace{-1.5em}
    \caption{
    Additional qualitative results for airplanes and lamps, showcasing accurate reconstruction of aircraft components and intricate lamp structures.
    }
    \label{fig:supp9}
    \vspace{-1.em}
\end{figure*}

Additional qualitative results across different object categories, including chair, airplane, and lamp, are provided to further support our experimental findings. For chair, further results can be found in Fig.~\ref{fig:supp7} and Fig.~\ref{fig:supp8}, where our model successfully preserves key structural elements such as detailed leg structure, backrest curvature, and armrest presence, even in abstract sketches. In the case of airplane, additional results are shown in Fig.~\ref{fig:supp9}, demonstrating that PASTA accurately generates aircraft components such as wings, fuselage, engines, and tail sections, with a high level of structural coherence. For lamp, additional 3D shape generations are also presented in Fig.~\ref{fig:supp9}, showcasing how our model captures intricate details such as delicate frame structures, lampshades, and support bases, producing models that closely align with input sketches. These qualitative results confirm that our method outperforms existing approaches across multiple object categories, offering superior generalization and the ability to handle complex and varied shapes with high fidelity and structural consistency.
\clearpage

\end{document}